\documentclass[journal,10pt,letterpaper]{IEEEtran}

\usepackage{cite}

\usepackage{graphicx}

\usepackage{amsmath}
\usepackage{amsthm}

\usepackage{amssymb}
\usepackage{bm}
\usepackage{mathtools}

\usepackage{booktabs}
\usepackage{nicefrac}
\usepackage{multirow}
\usepackage{threeparttable}
\usepackage{tabularx}

\usepackage{xcolor}
\usepackage{enumitem}
\usepackage{soul}
\usepackage{pifont}
\usepackage{url}
\usepackage[hidelinks]{hyperref}
\hypersetup{ pdftitle={Demistifying Data and Simulator Assumptions in Supervised Causal
Discovery}, pdfauthor={Pingchuan Ma, Rui
  Ding, Bojun Huang, Shuai Wang}, pdfkeywords={Causal discovery, Bayesian
  networks, supervised learning, foundation models, identifiability} }

\usepackage{tikz}
\usetikzlibrary{arrows.meta,positioning,fit,calc}
\tikzset{
  card/.style={draw=#1!60!black!90, line width=0.6pt, rounded corners=2.5pt,
      fill=#1!12, align=center, inner sep=3.5pt},
  chip/.style={fill=#1!18, rounded corners=4pt, inner sep=3pt,
      align=center, font=\scriptsize\bfseries, text=#1!45!black},
  badge/.style={circle, fill=#1!80!black, text=white, inner sep=0.6pt,
      minimum size=9pt, font=\tiny\bfseries},
  gnode/.style={circle, draw=black!75, line width=0.6pt, fill=white,
      minimum size=14pt, inner sep=0.5pt, font=\scriptsize},
  flow/.style={-{Latex[length=2.2mm, width=1.7mm]}, line width=0.8pt,
      draw=black!70},
  soft/.style={-{Latex[length=1.9mm, width=1.5mm]}, line width=0.6pt,
      draw=black!45, dashed},
  gedge/.style={line width=0.7pt, draw=black!75},
  gdir/.style={gedge, -{Latex[length=1.8mm, width=1.4mm]}},
  note/.style={font=\scriptsize, text=black!60, align=center},
}

\theoremstyle{definition}
\newtheorem{definition}{Definition}[]
\newtheorem{theorem}{Theorem}
\newtheorem{proposition}{Proposition}

\theoremstyle{definition}
\newtheorem*{remarkinner}{Remark}

\definecolor{vegaBlue}{HTML}{1f77b4}
\definecolor{vegaOrange}{HTML}{ff7f0e}
\definecolor{vegaGreen}{HTML}{2ca02c}

\newcommand{\parh}[1]{\smallskip\noindent\textbf{#1}}

\newcommand{\F}{Fig.}

\newcommand{\T}{Table}
\renewcommand{\S}{Sec.}

\newcommand{\ignore}[1]{}
\providecommand{\Description}[1]{}

\newcommand{\ci}{\perp\!\!\!\perp}
\newcommand{\dsep}{\perp\!\!\!\perp_G}

\usepackage{xspace}

\begin{document}

\title{Demistifying Data and Simulator Assumptions in Supervised Causal
Discovery}

\author{Pingchuan~Ma, Rui~Ding, Bojun~Huang, and~Shuai~Wang%
\thanks{P. Ma is with Zhejiang University of Technology, China.
E-mail: pma@zjut.edu.cn}%
\thanks{R. Ding is with Microsoft Research Asia.
E-mail: juding@microsoft.com}%
\thanks{B. Huang is with Sony Research.
E-mail: bojhuang@sony.com}%
\thanks{S. Wang is with HKUST, Hong Kong SAR.
E-mail: shuaiw@cse.ust.hk}}

\maketitle

\begin{abstract}
Supervised causal discovery learns to infer causal structure for a new
dataset from training datasets paired with structural labels. These training
pairs are typically simulated, making the simulator both a source of
supervision and a carrier of assumptions about causal graphs, mechanisms,
and noise. Understanding the resulting predictions therefore requires
examining how these assumptions supplement the information available in
observational data, which may be compatible with multiple causal graphs.
This paper examines that relationship across representative methods
available through June 2026. We organize these methods by prediction target,
prediction granularity, encoder, structural decoder, and training regime to
relate what each method predicts to how it uses data and simulator-based
supervision. Using this framework, we distinguish two questions: whether
the target is identifiable under the assumed model class, and whether a
trained predictor generalizes beyond its training distribution. Restrictions
on mechanisms and noise can make otherwise ambiguous causal directions
identifiable, but predictive accuracy under those restrictions does not
establish transfer when they change. This distinction motivates evaluation
that matches metrics to the identifiable graph target and tests changes in
graphs, mechanisms, and noise between training and deployment. Extending
such evaluation to real data also requires documenting the external causal
evidence and uncertainty behind benchmark reference graphs. Together, these
analyses guide method comparison and identify open questions in transfer,
test-time adaptation, and uncertainty assessment.
\end{abstract}


\IEEEpeerreviewmaketitle

\section{Introduction}
\label{sec:intro}

Causal discovery uses data to infer relations between variables that can
support reasoning about interventions. These relations are often represented
by a directed acyclic graph (DAG), with edges denoting direct causal
effects~\cite{pearl2009causality,peters2017elements}. Applications include
biology, epidemiology, and economics, where an intervention may be costly or
impossible and observational data are more readily
available~\cite{sachs2005causal,hernan2010causal,imbens2015causal}.

The available data may not determine a unique causal graph. A target is
\emph{identifiable} within a model class if any two models that generate the
same observed distribution agree on that target. Under the Markov and
faithfulness assumptions and in the absence of unobserved confounding,
conditional independencies identify a Markov equivalence class, represented
by a completed partially directed acyclic graph
(CPDAG)~\cite{spirtes2000causation}. PC and score-based GES recover this
CPDAG under their respective statistical conditions~\cite{spirtes2000causation,
chickering2002optimal}. FCI allows latent confounding and targets a partial
ancestral graph (PAG)~\cite{zhang2008completeness}. Stronger restrictions on
mechanisms or noise, such as those used by LiNGAM, can identify directions
left unresolved by conditional independence~\cite{shimizu2006linear,
hoyer2008nonlinear}. Each procedure must also estimate the relevant
statistical information from a finite sample.

\begin{figure*}[t]
\centering
\includegraphics[width=\textwidth]{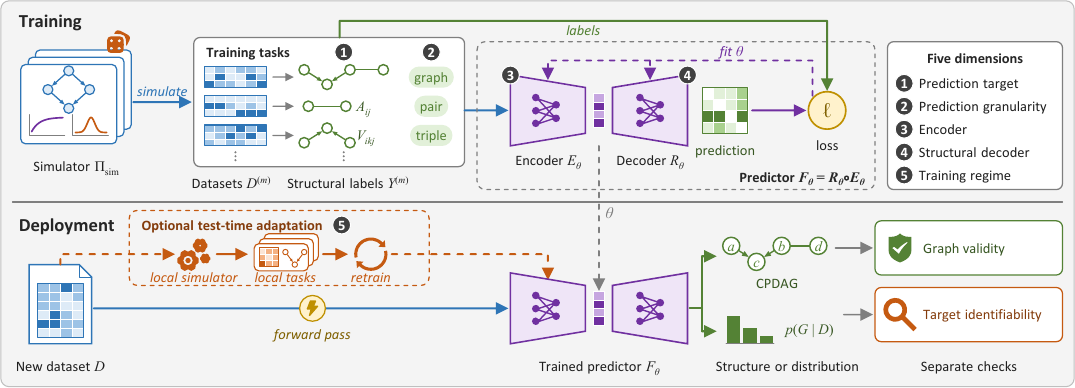}
\caption{A common SCD workflow. Training samples causal models, generates paired datasets and structural labels, and fits an encoder $E_\theta$ and decoder $R_\theta$. The prediction target and granularity specify whether labels describe complete graphs or local units, such as pairs and triples. Deployment encodes a new dataset and decodes a structure or a distribution over structures; the illustrated output is a CPDAG. Optional test-time adaptation fits a local simulator to the new dataset, generates local tasks, and retrains the predictor. Graph validity and target identifiability require separate checks. Numbered badges mark the five dimensions used to compare methods.}
\Description{A training band shows a simulator, datasets paired with graph-level and local structural labels, and an encoder and decoder fitted with a loss. A deployment band applies the trained predictor to a new dataset, with an optional test-time adaptation loop, and outputs a CPDAG or a distribution over graphs, followed by graph validity and target identifiability checks.}
\label{fig:taxonomy-pipeline}
\end{figure*}

Supervised causal discovery (SCD) trains a predictor whose input is a dataset
and whose output describes causal structure. Some studies call this task
supervised causal learning (SCL)~\cite{zhanglearning,deng2026ttt}. We use SCD
throughout and retain published method names such as TTT-SCL. For training,
a simulator typically samples many causal models, generates a dataset from each
model, and pairs the dataset with a structural
label~\cite{lopez2015towards,li2020supervised}.
The simulator supplies labels that are rarely available for real observational
datasets. A trained predictor then estimates structure for a new dataset.
Some methods supplement these training pairs with real examples, external
knowledge, or estimates from classical discovery algorithms.
\F~\ref{fig:taxonomy-pipeline} summarizes this workflow.

SCD methods differ in what they predict and how they use the input dataset.
A predictor may directly estimate a
structure, such as a skeleton, a set of v-structures, a CPDAG, or a
DAG~\cite{ma2022ml4s,dai2023ml4c,zhanglearning,ke2022learning}. It may estimate
edge probabilities or a distribution over graphs, from which graph estimates
or samples can be obtained~\cite{lorch2022amortized,dhirmeta}. Some methods,
such as SEA, use graph estimates computed on subsets of variables as inputs
to a predictor of the complete graph~\cite{wu2025sample}. Comparing these
methods requires distinguishing the final graph target from the local
relations used as inputs or training labels.

The model architecture determines how a dataset is converted into that target.
An encoder can summarize conditional-independence statistics, process samples
with attention, or incorporate knowledge and estimates from other
algorithms~\cite{ma2022ml4s,lorch2022amortized,xu2026knowledge,wu2025sample}.
A decoder converts this representation into local scores or a graph. Local
scores may require aggregation and orientation rules; other decoders construct
a DAG through an ordering of the variables~\cite{dhirmeta,thompson2026arrow}.
These decoding procedures determine whether the predicted graph is valid.
Identifying its edge directions also requires sufficient information in the
data under the assumed causal model.

Training can occur before deployment, on the test dataset, or at both stages.
Pretrained predictors can reuse computation across datasets and reduce the
cost of inference to a forward pass or a sampling procedure. The total saving
depends on the cost of pretraining and the number of subsequent
uses. Methods that fit a local simulator or retrain a predictor for each test
dataset, including ML4S, TTT-SCL, and TICL, incur additional search,
simulation, and optimization costs at deployment~\cite{ma2022ml4s,deng2026ttt,chen2026ticl}.
SPOT instead combines a pretrained skeleton predictor with per-dataset graph
optimization~\cite{ma2024scalable}. Broad
pretraining and test-time adaptation address different sources of mismatch between
training tasks and the dataset being analyzed, and can be combined.

The identifiability results used in classical discovery also apply to SCD.
For example, a supervised method under the same observational assumptions as
PC can target the same CPDAG. Its predictions also depend on the training
distribution. A simulator may favor one member of a Markov equivalence class,
so training on full-DAG labels can favor directions that are not identified
by the observed distribution~\cite{montagna2024demystifying,zhanglearning}.
A posterior can also favor one compatible graph because of its prior.
Even a predictor trained on CPDAG labels can perform poorly on unfamiliar mechanisms or graph
patterns. Target identifiability and generalization across training and test
distributions therefore require separate checks.

Predictions can also help investigate the assumptions encoded in a simulator.
For instance, varying the allowed graphs or mechanism family and examining
predicted edge probabilities at different sample sizes can help determine
which simulator assumptions affect predictions. These experiments also need
to assess calibration and sensitivity to the prior, because a predictor can
be confident even when the data do not identify a unique graph.

\parh{Scope and organization.}~This paper examines the assumptions and model
choices at each step of an SCD workflow. After reviewing identifiability, we
describe how training tasks and labels are produced, how the input dataset
is encoded, and how predictions
are assembled into a graph. We use five corresponding dimensions to compare
methods: prediction target, prediction granularity, encoder, structural
decoder, and training regime. For each target, we state the assumptions needed
to interpret it causally. This organization lets readers compare methods that
estimate the same target and identify differences in their inputs, models, and
training procedures.

The paper provides three analyses:
\begin{itemize}
  \item We relate the targets of representative SCD methods to classical
  identifiability results and describe their encoders, decoders, and training
  regimes within a common workflow.
  \item We examine how simulators generate structural labels, directional
  asymmetries, and other distributional constraints, and where predictions
  based on those constraints may fail on new data.
  \item We discuss evaluation with metrics matched to graph targets, separate
  training and test simulators, and causal reference graphs supported by external
  evidence. These considerations also apply to broad pretraining and
  adaptation to individual test datasets.
\end{itemize}

\S~\ref{sec:background} introduces structural causal models and
identifiability. \S~\ref{sec:implications-learning} presents the SCD workflow
and compares representative methods. \S~\ref{sec:simulators} examines
simulator assumptions, transfer to new datasets, and benchmark design.
\S~\ref{sec:conclusion} discusses the remaining open questions.

\section{Background}
\label{sec:background}

This section defines the causal models and graph targets used in this paper.
We distinguish assumptions about the complete causal system from assumptions
about which variables are observed, then state what conditional independence
and functional restrictions can identify.

\subsection{Conceptual View of Causality}

Causal discovery concerns relations between variables in a causal
model. It is distinct from \emph{actual causality}, which asks whether a
particular event caused another in a specific context. Lewis's counterfactual
account analyzes dependence between events~\cite{lewis1973causation}, but
simple but-for dependence does not cover cases such as preemption. Structural
accounts, including the Halpern--Pearl definition, address actual causality by
considering interventions under specified
contingencies~\cite{halpern2005causes}. We use structural causal models to
define mechanisms and interventions; a definition of actual causality is not
needed to state the graph discovery task.

\begin{figure*}[t]
\centering
\includegraphics[width=\textwidth]{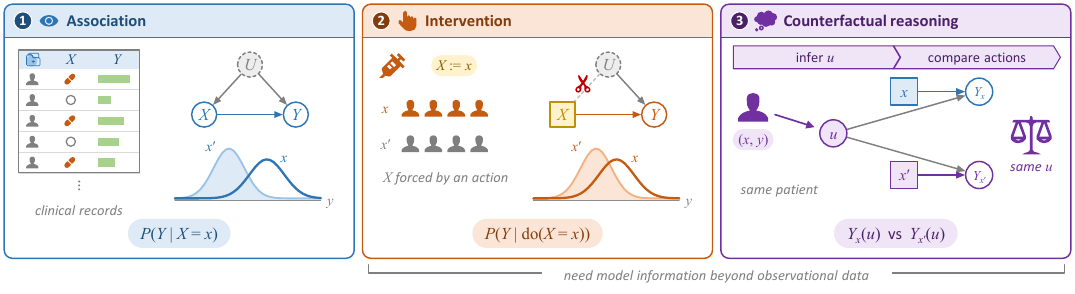}
\caption{Three causal queries, illustrated by treatment $X$ and recovery $Y$. Clinical records describe association; intervening replaces the treatment mechanism by $X=x$. Counterfactual reasoning uses a causal model to infer an exogenous context $u$ and compare actions for the same patient, $Y_x(u)$ versus $Y_{x'}(u)$. The table and curves are schematic. Observational data alone need not identify the model information required by the latter two queries.}
\Description{Three panels show clinical records with a confounded graph and conditional densities, a randomized intervention that cuts the edge into X, and a twin network in which one patient's context u is shared by two actions.}
\label{fig:conceptual-causality}
\end{figure*}

Pearl's ladder of causality distinguishes three kinds of query~\cite{pearl2009causality,bareinboim2022pearl}. The first
rung is \emph{association}, which concerns observational quantities such as
$P(Y\mid X=x)$ and answers questions about what is observed together. The second
rung is \emph{intervention}, which concerns quantities such as
$P(Y\mid\operatorname{do}(X=x))$ and asks what would happen if an action forced
$X$ to take a chosen value. The third rung is \emph{counterfactual reasoning},
which asks what would have happened to the same unit under a different
condition. \F~\ref{fig:conceptual-causality} illustrates the three rungs with
treatment $X$ and recovery $Y$. Causal discovery often uses observational data
to estimate a structure that will be used for interventional or counterfactual
reasoning.
The distinction between these queries explains why this use requires
assumptions beyond the observed associations.

Structural causal models define these queries through equations for variable
generation and intervention. We next introduce these models and their graphs,
then define causal discovery as the recovery of a graph or equivalence class
under stated assumptions.

\subsection{Notation, Structural Causal Models, and Graphical Models}

We use $X$ to denote a scalar variable, $\bm{X}$ to denote a set of variables
and $P_{\bm{X}}$ to denote the joint distribution over $\bm{X}$. Each variable
corresponds to a node in a graph. A directed edge $X \to Y$ implies that $X$ is
a direct cause (or parent) of $Y$. The graph $G = (V_G, E_G)$ represents a
causal structure, where $V_G$ is the set of variables and $E_G$ is the set of
edges. $\operatorname{Pa}_G(X) \coloneqq \{X' \mid X' \to X \in E_G\}$ denotes
the parents of $X$ in $G$.

We use the framework of a \emph{structural causal model}
(SCM)~\cite{pearl2009causality} to connect this graph to a data-generating
process. An SCM associates each variable $X_i \in \bm{X}$ with a structural
equation
\begin{equation}
    X_i = f_i(\operatorname{Pa}_G(X_i), N_i),
\end{equation}
where $f_i$ is a deterministic mechanism, $\operatorname{Pa}_G(X_i)$ are the
parents of $X_i$ in $G$, and $N_i$ is an exogenous noise term. The model also
specifies a joint distribution for the noise variables. Mutually independent
noise terms are a standard assumption when shared causes are included
explicitly in the graph. After unobserved common causes are marginalized out,
the effective noise terms for observed variables may be dependent. An SCM
therefore need not have independent errors on the observed variables.
The graph records which variables enter each mechanism directly. Within the
SCM, an intervention replaces a structural equation. Applying this model to a
real system requires assessing whether its equations and intervention
assumptions describe that system.

With latent confounding, an observed margin can be represented by mixed
graphs. In a maximal ancestral graph (MAG), directed and bidirected edges
summarize ancestral relations and dependence paths through omitted variables;
a bidirected edge need not correspond to a single hidden common cause.
Partial ancestral graphs (PAGs) represent Markov equivalence classes of
MAGs~\cite{richardson2002ancestral,zhang2008completeness}. We focus on DAGs and
CPDAGs under causal sufficiency, and MAGs/PAGs without selection bias when
latent confounding is allowed.

\subsection{Task Description}

Causal discovery asks which features of a causal model can be recovered from
the available data under a stated model class. Let $\eta$ specify a causal
model with graph $G$, let $P_\eta^{\mathrm{obs}}$ denote its observed
distribution, and let $\mathcal T(G)$ be a graph target. The target is
identifiable in a class $\mathfrak M$ if
\begin{equation}
\begin{gathered}
 P_\eta^{\mathrm{obs}}=P_{\eta'}^{\mathrm{obs}}
 \quad\Longrightarrow\quad
 \mathcal T(G)=\mathcal T(G'),\\
 \eta,\eta'\in\mathfrak M.
\end{gathered}
\label{eq:target-identifiability}
\end{equation}
The target may be a skeleton, v-structures, an equivalence class, or a full
graph. Interventions and background knowledge can restrict the compatible
models further. Identifiability concerns the population distribution;
estimation from finite data introduces additional uncertainty.

An estimator takes a dataset $D$ to a prediction
$\widehat{\mathcal T}=\mathcal F(D)$. It may use independence tests,
score-based search, or a learned function. A Bayesian method instead specifies
a prior and estimates a posterior such as $p(G\mid D)$. This posterior is a
distribution over candidate graphs conditional on the data and prior, rather
than a graph feature $\mathcal T(G)$; it can remain non-degenerate even with
unlimited observational data. The following assumptions describe the graph
features that these procedures can identify.

\subsection{Standard Assumptions}

Let $\bm V=\bm X\cup\bm H$ contain the endogenous variables of an acyclic
causal model, with observed variables $\bm X$ and latent variables $\bm H$.
We first state A1 and A2 for the full DAG $G$ over $\bm V$ and its joint
distribution $P_{\bm V}$. A3 then states when the observed variables contain all their
common causes. Under A3, the relevant observed causal model admits a DAG
representation over $\bm X$, so A1 and A2 can be used there without latent
common causes.

\paragraph{A1: Markov Assumption}
A directed acyclic graph (DAG) $G$ satisfies the Markov condition with respect
to $P_{\bm V}$ if:

\begin{equation}
    P_{\bm V} = \prod_{X \in \bm V} P(X \mid \operatorname{Pa}_G(X)).
\end{equation}

This factorization implies that each variable is conditionally independent of
its non-descendants given its parents. Equivalently, the global Markov property
states:

\begin{equation}
    X \dsep Y \mid \bm{Z} \Rightarrow X \ci Y \mid \bm{Z},
\end{equation}

where $\dsep$ denotes d-separation in the graph and $\ci$ denotes conditional
independence in the distribution. D-separation is defined as follows.
\begin{definition}[d-Separation]
Two nodes $X$ and $Y$ are d-separated by a set of nodes $\bm{Z}$ in $G$ if every
undirected path between $X$ and $Y$ is blocked by $\bm{Z}$. A path is blocked
if at least one of the following holds:
\begin{itemize}
    \item Some non-collider node $W$ on the path satisfies $W \in \bm{Z}$.
    \item Some collider node $W$ on the path satisfies neither $W \in \bm{Z}$
    nor ``a descendant of $W$ is in $\bm{Z}$.''
\end{itemize}
\end{definition}

\paragraph{A2: Faithfulness.}  
The full joint distribution $P_{\bm V}$ is faithful to a DAG $G$ if all conditional independencies in
$P_{\bm V}$ correspond to d-separations in $G$:

\begin{equation}
    X \ci Y \mid \bm{Z} \Rightarrow X \dsep Y \mid \bm{Z}.
\end{equation}

Together, the Markov and faithfulness assumptions imply an equivalence between
statistical independencies and graphical constraints in the DAG. This
equivalence supports causal discovery methods that use conditional
independence tests to infer the structure of $G$.

\paragraph{A3: Causal Sufficiency.}  
Causal sufficiency assumes that all common causes of the observed variables are
themselves observed.

A3 is a restriction on what is observed, not a consequence of A1. For example,
a full DAG $U\to X$, $U\to Y$ can be Markov and faithful while the observations
of $X,Y$ omit $U$. Marginalizing $U$ generally invalidates the factorization
associated with the induced observed subgraph, which has no edges. An observed
distribution may nevertheless be Markov to some other DAG; that statistical
factorization alone does not establish causal sufficiency. We state A3
explicitly when a result concerns the causal DAG on observed variables.

\subsection{Identifiability from Conditional Independence}

Under a given set of assumptions, which features of the causal graph can be
recovered from observational data? Under A1--A3, conditional independencies
identify the skeleton and unshielded colliders of the observed causal DAG.
Together these determine its Markov equivalence class. They need not determine
every edge direction.

\begin{definition}[Skeleton]
The skeleton of a DAG $G$ is an undirected graph $S = (V_S, E_S)$ with
$V_S=V_G$ such that:
\begin{equation}
    (X \to Y \in E_G \text{ or } Y \to X \in E_G) \iff (X - Y \in E_S).
\end{equation}
\end{definition}

\begin{theorem}[Spirtes et al.~\cite{spirtes2000causation}]
\label{thm:constraint}
Let $G$ be the causal DAG over the observed variables $\bm X$. Under
A1--A3, distinct $X$ and $Y$ are adjacent in $G$ if and only if there exists no
subset $\bm{Z} \subseteq \bm{X} \setminus \{X, Y\}$ such that $X \ci Y \mid
\bm{Z}$.
\end{theorem}

The theorem concerns the causal DAG over the observed variables $\bm X$ and
therefore requires A3, although the graphical definition of a skeleton does
not. Under A1--A3, the observed causal skeleton is uniquely identifiable from
all conditional independence (CI) relations. Learning the skeleton alone is
insufficient for determining causal directions. Orientation rules are used to
infer edge directions, typically starting with \emph{unshielded triples} (i.e.,
$X-Z-Y$ where $X$ and $Y$ are not adjacent).

\begin{proposition}[Verma and Pearl~\cite{verma2022equivalence}]
\label{prop:immorality}
Under A1--A3, let $X-Z-Y$ be an unshielded triple in the skeleton of the
observed causal DAG. It is a v-structure (or immorality), $X\to Z\leftarrow Y$,
if and only if $Z$ belongs to none of the sets
$\bm S\subseteq\bm X\setminus\{X,Y\}$ for which $X\ci Y\mid\bm S$.
\end{proposition}

Applying this rule identifies \emph{v-structures}. The remaining edges are then
oriented using rules such as Meek's rules~\cite{meek1995causal}, which enforce
acyclicity and propagate orientations. However, under canonical assumptions,
only the \textit{Markov equivalence class} is identifiable.

\begin{definition}[Markov Equivalence Class~\cite{peters2017elements}]
Two DAGs $G_1$ and $G_2$ are Markov equivalent if they imply the same set of CI
relations:
\begin{equation}
    \mathcal{M}(G_1) = \mathcal{M}(G_2).
\end{equation}
where $\mathcal{M}(G)$ is the set of distributions that are Markovian with
respect to $G$.
\end{definition}

\begin{proposition}
\label{prop:mec}
Two DAGs are Markov equivalent if and only if they have the same skeleton and
the same set of immoralities.
\end{proposition}

Under A1--A3 and access to correct CI relations, these results and complete
orientation rules recover the CPDAG~\cite{spirtes2000causation,meek1995causal}.
A DAG is recovered uniquely only when this equivalence class has one member.
With latent confounding, the full causal DAG still exists, but observed CI
relations generally identify an equivalence class of MAGs represented by a
PAG. This requires the appropriate marginal graphical model and orientation
rules~\cite{richardson2002ancestral,zhang2008completeness}.

\subsection{Additional Assumptions}

Directions unresolved within a Markov equivalence class require information
beyond its CI relations. One source is a restriction on functional form and
noise. We use A4 to refer to functional model classes whose restrictions
ensure identifiability.
Sufficiently informative interventions or background knowledge provide other
sources of orientation information.

\paragraph{A4: Identifiable Functional Model Classes (IFMOCs)}
\emph{Identifiable Functional Model Classes}
(IFMOCs)~\cite{peters2011identifiability} restrict each structural equation
to a function of the variable's parents and an independent noise term.
Additional conditions on the functions or noise distributions ensure that
the causal directions are identifiable.

\paragraph{A4.1: Additive Noise Models (ANM)}  
In additive noise models, each variable is assumed to be generated as
\begin{equation}
    X_i = f_i(\operatorname{Pa}_G(X_i)) + N_i,
\end{equation}
where $f_i$ is a linear or nonlinear deterministic function and the noise
terms are mutually independent. In the bivariate case $Y=f(X)+N_Y$ with
$N_Y\ci X$, the reverse representation $X=g(Y)+N_X$ with $N_X\ci Y$
generally does not exist within identifiable ANM classes. This asymmetry
yields identifiability of the causal direction
under the regularity and non-degeneracy conditions of the relevant ANM
identifiability result~\cite{hoyer2008nonlinear,peters2014causal}. Nonlinearity
alone is not sufficient in every case. Linear Gaussian models with unrestricted
noise variances are a familiar exception; heteroscedastic models require a
separate analysis because their noise scale depends on the parents.

Several practical methods build upon this assumption:
\begin{itemize}
    \item The \textbf{ANM method}~\cite{peters2014causal} tests for independence
    between residuals and predictors in both directions and selects the one
    consistent with the ANM assumption.
    \item \textbf{LiNGAM} (Linear Non-Gaussian Acyclic
    Model)~\cite{shimizu2006linear} specializes ANM to linear functions
    and exploits non-Gaussian noise to identify the full DAG. The original
    estimator uses independent component analysis; DirectLiNGAM estimates
    a causal ordering through successive regressions and independence
    comparisons~\cite{shimizu2011directlingam}.
\end{itemize}

\begin{figure}
    \centering
    \includegraphics[width=0.95\columnwidth]{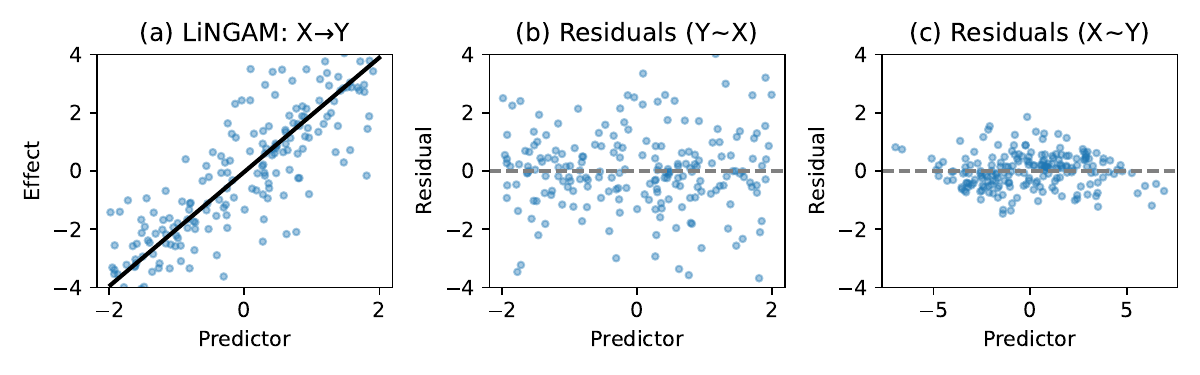}
\caption{LiNGAM asymmetry illustrated by 200 samples with
$X\sim\mathcal U[-2,2]$, $Y=2X+\varepsilon$, and independent
$\varepsilon\sim\mathrm{Lap}(0,1)$. The population forward noise is independent
of $X$, whereas the population reverse linear-regression residual depends
on $Y$. The plots show fitted finite-sample residuals and illustrate this
asymmetry; they are not independence tests.}

    \label{fig:anm}
\end{figure}

\F~\ref{fig:anm} illustrates the independent-noise asymmetry in a linear
non-Gaussian model. Population noise independence is a model property;
estimated regression residuals only approximate that noise. In this example,
the LiNGAM assumptions identify $X\to Y$. Dependence after reverse linear
regression alone would not rule out every nonlinear reverse model.

\paragraph{A4.2: Post-Nonlinear Models (PNL)}  
Post-nonlinear models generalize ANMs by allowing a nonlinear distortion of the
additive functional form~\cite{zhang2009identifiability}. Each variable is
assumed to follow
\begin{equation}
    X_i = g_i\!\left(f_i(\operatorname{Pa}_G(X_i)) + N_i\right),
\end{equation}
where $f_i$ is a deterministic function of the parents, $N_i$ is an independent
noise variable, and $g_i$ is an invertible nonlinear function. The outer
nonlinearity $g_i$ models effects such as sensor distortions or nonlinear
transformations applied to the observed signal. This formulation also induces
asymmetries: for the true causal direction, the noise remains independent after
applying the inverse of $g_i$, whereas the reverse direction fails to satisfy
this property in general. 

\begin{figure*}[t]
\centering
\includegraphics[width=\textwidth]{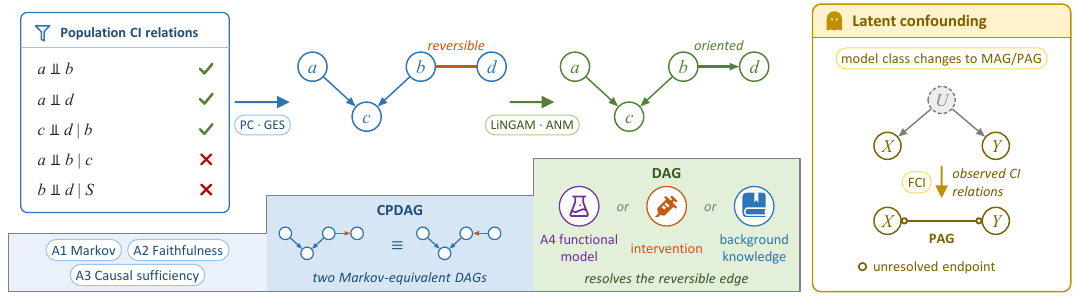}
\caption{Graph targets under different assumptions. In an observed DAG satisfying A1--A3, population CI relations identify a CPDAG. The example $a\to c\leftarrow b$ with $b-d$ has one reversible edge. Additional direction information can resolve it. Allowing latent confounding changes the model class to MAG/PAG semantics; circles in the two-variable PAG indicate unresolved endpoints. Chips on the arrows name representative algorithms.}
\Description{A staircase rises from assumptions A1 to A3 and population CI relations to a CPDAG with two Markov-equivalent DAGs, and then to a DAG oriented by an A4 functional model, an intervention, or background knowledge. A separate card shows latent confounding and a PAG with circle endpoints.}
\label{fig:identifiability-ladder}
\end{figure*}

Identifiable functional classes can resolve directions left open by CI
relations when their additional conditions hold. Neural parameterizations can
approximate structural functions, but flexibility and acyclicity alone do not
preserve these identifiability conditions.
\F~\ref{fig:identifiability-ladder} separates the observed DAG setting from
the setting with latent confounding. Allowing latent variables changes the
model class; it is not an intermediate step toward identifying a full DAG.

\subsection{Classical Causal Discovery Methods}

Classical causal discovery methods solve a discovery problem separately for
each dataset. They use conditional independence tests, graph scores, or
continuous optimization. Although their algorithms differ, the causal
interpretation of their output depends on the assumptions that identify the
graph target.

\parh{Constraint-based methods.}~Constraint-based algorithms recover graph
structure by testing conditional independence relations and translating them
into graphical constraints. The PC algorithm is the canonical example under
causal sufficiency: it removes adjacencies when a separating set is found and
then orients v-structures and compelled edges using orientation rules such as
Meek's rules~\cite{spirtes2000causation,meek1995causal}. When latent
confounders or selection effects are allowed, FCI-style methods replace DAG or
CPDAG semantics with ancestral-graph or PAG semantics, so that uncertain
endpoints and possible hidden common causes are represented
explicitly~\cite{richardson2002ancestral,colombo2012learning,ogarrio2016hybrid}.

CI tests such as the kernel-based conditional-independence test can be used
within PC~\cite{zhang2011kernel}. Each edge deletion or orientation follows
from CI relations under the
algorithm's assumptions. However, CI tests can be unreliable with finite
samples. Even with correct CI relations, these methods generally recover an
equivalence class unless additional information identifies the remaining
directions.

\parh{Score-based methods.}~Score-based algorithms search over graphs using a
criterion that trades off data fit and model complexity. Greedy Equivalence
Search (GES) is representative: it searches over equivalence classes and, under
standard score assumptions, targets the CPDAG rather than an arbitrary DAG
representative~\cite{chickering2002optimal}. Hybrid methods such as MMHC first
use constraint-based screening to restrict the candidate neighborhood and then
apply score-based search, improving scalability while retaining an explicit
graphical search objective~\cite{tsamardinos2006max}. Exact and improved search
methods further study score optimization and the assumptions needed for
recovery~\cite{ng2021reliable}.

Compared with
constraint-based methods, score-based methods can be more robust to noisy CI
decisions, but they inherit the assumptions encoded by the score, likelihood
family, sparsity penalty, and search space. Thus, a high-scoring DAG should not
be interpreted as fully causally identified unless the corresponding
identifiability assumptions justify that target.

\parh{Differentiable methods.}~Differentiable causal discovery casts structure
learning as continuous optimization by parameterizing an adjacency matrix and
enforcing acyclicity through a smooth constraint or penalty. NOTEARS is the
prototype of this line: it replaces discrete DAG search with a continuous
objective subject to an analytic acyclicity
constraint~\cite{zheng2018dags}. Subsequent methods extend this idea with nonlinear
parameterizations, neural graph models, alternative DAG constraints, and more
stable optimization strategies, including DAG-GNN, nonparametric NOTEARS,
DAGMA, and related variants~\cite{yu2019dag,zheng2020learning,bello2022dagma,
wei2020dags,nazaret2024stable}. These formulations allow gradient-based
optimization of the graph parameters.

An exactly satisfied acyclicity constraint ensures that the result is a DAG;
numerical penalties and approximate optimization require a separate validity
check. Neither
acyclicity nor a low reconstruction loss establishes that its directions are
identifiable. That requires the statistical conditions discussed above, such
as A1--A3 for a CPDAG or the additional restrictions of an A4 functional model
for a full DAG.

SCD reuses computation by training a predictor across datasets. The next
section describes how these predictors are trained and how their outputs
relate to the graph targets and assumptions introduced here.

\section{Supervised Causal Discovery: Workflow and Method Comparison}
\label{sec:implications-learning}

The background results specify which graph targets are identifiable under
given assumptions. SCD estimates these targets with a predictor trained on
datasets paired with structural labels. Each training task contains a complete
dataset, even when the predictor estimates individual edges or triples.
\F~\ref{fig:taxonomy-pipeline} shows how simulation, encoding, decoding,
and training form an SCD workflow.

\parh{Literature selection.}~Existing surveys discuss graphical causal
discovery~\cite{glymour2019review}
and continuous optimization~\cite{vowels2022d}; this paper focuses on
learning prediction rules that map datasets to causal structure. We selected
representative studies primarily through Google~Scholar, including publications
and preprints available through June 2026. We include methods trained on
datasets paired with structural labels, either across tasks or at test time,
and ADAG as a related approach without supervised graph labels. Work limited
to causal effect estimation, time-series discovery, or causal representation
learning from unstructured inputs is outside the comparison.

\subsection{From Simulated Tasks to Graph Predictions}
\label{sec:scl-workflow}

A simulator first samples a causal model, generates a dataset $D^{(m)}$, and
computes a structural label $Y^{(m)}$ from the sampled graph. For example, to
train a CPDAG predictor, we can simulate data from a DAG and convert that DAG
to its CPDAG for supervision. If the model predicts local units, the same task
provides adjacency labels and collider labels for its unshielded triples.
This choice determines the \emph{prediction target} and \emph{prediction
granularity}. Full-DAG labels are also available from the simulator, but
whether their orientations can be inferred from the data depends on the model
class, as discussed in \S~\ref{sec:background}.

The predictor encodes the dataset and decodes a prediction. Write
$F_\theta=R_\theta\circ E_\theta$, where $E_\theta$ is an encoder and
$R_\theta$ includes the prediction head and any graph assembly procedure.
A typical training objective is
\begin{equation}
 \widehat\theta\in\arg\min_\theta\frac{1}{M}\sum_{m=1}^M
 \ell\!\left(R_\theta(E_\theta(D^{(m)})),Y^{(m)}\right).
 \label{eq:scl-workflow}
\end{equation}
For the CPDAG example, the encoder may extract CI statistics or learn a
representation of the samples. The prediction heads estimate adjacencies and
colliders, and orientation rules assemble them into a graph. A posterior
predictor instead estimates a conditional graph distribution or its marginals
using an appropriate probabilistic loss. Some methods also use knowledge or
estimates on variable subsets as encoder inputs.

At deployment, a fixed predictor applies the learned rule to a new dataset.
Its cost includes any feature extraction, classical estimators, graph
assembly, or sampling required by that rule. An adaptive method generates or
reweights training tasks for the new dataset and fits the predictor again.
The \emph{training regime} specifies both the source of training tasks and
when this fitting occurs. These stages give the five dimensions used in
\T~\ref{tab:unified-taxonomy}: prediction target, prediction granularity,
encoder, structural decoder, and training regime. The target column records
what each method predicts and the assumptions relevant to its interpretation.

Simulators are used by local methods such as RCC and ML4S, posterior methods
such as AVICI and BCNP, and broadly pretrained methods such as
Arrow~\cite{lopez2015towards,ma2022ml4s,lorch2022amortized,dhirmeta,
thompson2026arrow}. Their role is to produce paired tasks with known
structure. Classical estimators can provide intermediate inputs, as in SEA,
and external knowledge can supplement the data, as in
Kode~\cite{wu2025sample,xu2026knowledge}. We also include ADAG as a related
approach that trains a reusable predictor by reconstruction across
domains, without supervised graph labels~\cite{yin2025learning}.
\S~\ref{sec:simulators} examines the distributions and constraints induced
by these training sources in more detail.

\begin{table*}[!t]
\centering
\caption{Representative methods compared by prediction target, granularity, model
components, and training regime. The assumptions qualify the interpretation of
the target. A simulator prior defines a prediction problem but does not by
itself guarantee identification. ADAG is included as a related reconstruction
approach without supervised graph labels.}
\label{tab:unified-taxonomy}
\small
\setlength{\tabcolsep}{2.8pt}
\renewcommand{\arraystretch}{1.12}
\begin{tabularx}{\textwidth}{@{}>{\raggedright\arraybackslash}p{0.16\textwidth} >{\raggedright\arraybackslash}p{0.14\textwidth} >{\raggedright\arraybackslash}p{0.13\textwidth} >{\raggedright\arraybackslash}p{0.12\textwidth} >{\raggedright\arraybackslash}p{0.15\textwidth} >{\raggedright\arraybackslash}X@{}}
\toprule
\textbf{Method} & \textbf{Target / Assumption} & \textbf{Granularity} & \textbf{Encoder} & \textbf{Decoder} & \textbf{Training Regime} \\
\midrule
\multicolumn{6}{@{}l}{\textit{Local and global supervised predictors}}\\
\addlinespace[1pt]
RCC~\cite{lopez2015towards} & Pairwise relation (sim. prior) & Bivariate pair & Kernel mean embedding & Classifier & Synthetic pairs; transductive tuning on T\"ubingen inputs \\
\addlinespace[2pt]
ML4C~\cite{dai2023ml4c} & CPDAG given skeleton (A1--A3) & Unshielded triple & Data + skeleton; vicinity statistics & Classifier + orientation rules & Synthetic triple labels \\
\addlinespace[2pt]
DAG-EQ~\cite{li2020supervised} & DAG (sim. prior) & DAG & Correlation + EQ network & Edge scores + cycle-avoiding assembly & Synthetic DAG tasks \\
\addlinespace[2pt]
CSIvA~\cite{ke2022learning} & DAG labels (sim. prior) & Graph & Transformer (alt.\ attn.) & Sequential; no DAG constraint & Observational / interventional tasks \\
\addlinespace[2pt]
SiCL~\cite{zhanglearning} & CPDAG (A1--A3) & CPDAG & Transformer (feature attn.) & CPDAG-aware decoder & Skeleton / collider labels \\
\midrule
\multicolumn{6}{@{}l}{\textit{Posterior SCD}}\\
\addlinespace[1pt]
AVICI~\cite{lorch2022amortized} & Directed-graph posterior approx. (sim. prior) & Edge marginals & Transformer (axial attn.) & Edge head; optional DAG penalty & SCM and dynamical GRN tasks \\
\addlinespace[2pt]
BCNP~\cite{dhirmeta} & DAG posterior (sim. prior) & DAG samples & Set transformer & Hardened perm.\ + triangular edges & Bayesian-network task prior \\
\midrule
\multicolumn{6}{@{}l}{\textit{Knowledge-informed SCD}}\\
\addlinespace[1pt]
Kode~\cite{xu2026knowledge} & DAG posterior (sim. prior + knowledge) & DAG samples & Data + knowledge & Knowledge-biased DAG sampler & Knowledge curriculum \\
\midrule
\multicolumn{6}{@{}l}{\textit{Supervised predictors combined with graph estimation}}\\
\addlinespace[1pt]
SPOT~\cite{ma2024scalable} & Skeleton posterior; downstream MAG estimate & Adjacencies, then graph & Local CI-test statistics & Classifier + MAG optimization & Synthetic pretraining; per-dataset graph fitting \\
\addlinespace[2pt]
SEA~\cite{wu2025sample} & DAG (sim. prior) & Subset inputs; global output & Stats + marginal-estimate encoder & Axial-attn aggregator & Sample-estimate-aggregate \\
\midrule
\multicolumn{6}{@{}l}{\textit{Broad pretraining and related predictors}}\\
\addlinespace[1pt]
ADAG~\cite{yin2025learning} & DAG (linear-additive) & Weighted DAG & Attention kernel map & Attn kernel + DAG constraint & Shared structure/order SEMs \\
\addlinespace[2pt]
Arrow~\cite{thompson2026arrow} & DAG (sim. prior) & DAG & Transformer (table) & Skeleton + total order & DAG labels; composite likelihood \\
\midrule
\multicolumn{6}{@{}l}{\textit{SCD with test-time adaptation}}\\
\addlinespace[1pt]
ML4S~\cite{ma2022ml4s} & DAG skeleton (A1--A3) & Edge adjacency & Local CI-test statistics & Cascade classifiers & Fitted pseudo BN; vicinal simulation and training \\
\addlinespace[2pt]
TTT-SCL~\cite{deng2026ttt} & Base SCD target (local sim. prior) & Base SCD units & Re-trained SCD encoder & Base SCD decoder & Test-aligned local simulator \\
\addlinespace[2pt]
TICL~\cite{chen2026ticl} & Equivalence class + intervention orientations & Skeleton + orientation & JCI edge/triplet features & PC-style two-phase decoder & Self-augmented test-time data \\

\bottomrule
\end{tabularx}
\end{table*}

The table groups related implementations for readability. These groups can
overlap: a posterior predictor can use broad pretraining, and a model using
external knowledge can also be adapted to a test dataset.

\subsection{Prediction Targets and Model Components}

\parh{Prediction target.}~The target is the quantity the predictor is trained to
estimate. Under A1--A3, a skeleton, unshielded colliders, and the resulting
CPDAG are identifiable from population CI relations. With latent confounding,
MAG/PAG targets require the corresponding graphical assumptions. For full-DAG
labels, a study must identify the additional orientation information or
interpret the output as a prediction conditional on the simulator prior.
Edge marginals and graph samples both describe uncertainty about a graph.
However, marginals alone do not specify the dependencies among edges that
can be represented by samples from a joint graph distribution.

\parh{Prediction granularity.}~Granularity specifies whether a prediction is
made for a pair, a triple, a variable subset, or the whole graph. In the
workflow example, a CPDAG is the final target while adjacencies and colliders
are the supervised units. RCC predicts bivariate directions, and SEA uses
estimates on variable subsets as inputs to a global predictor. Local units
provide many labels per simulated task and can reduce model size, but their
predictions must be combined into a consistent graph. Predicting a pairwise
direction requires different information from predicting a pairwise
adjacency, even though both use the same prediction granularity.

\parh{Encoder.}~The encoder maps the input dataset and any side information to
a representation. Local methods can use CI-test statistics; attention models
can learn from the samples directly. Other encoders also process background
knowledge about possible ancestral relations or graph estimates computed on
variable subsets. Comparing encoders requires examining which information
they retain, how their outputs change when variables or samples are reordered,
and how their computational costs grow with dataset size.

\parh{Structural decoder.}~The decoder turns the representation into scores,
graphs, or graph samples. Independent edge scores are easy to predict, but
thresholding them can introduce cycles or inconsistent collider decisions.
Orientation rules, graph post-processing, and decoders based on topological
orders address different parts of this problem. A CPDAG decoder must return a
valid equivalence-class representation; a DAG sampler must generate acyclic
graphs. These requirements ensure a valid graph representation. Whether the
data identify its directions depends on the causal assumptions.

\subsection{Training Data and Deployment}
\label{sec:training-regime}

\parh{Training regime.}~The fifth dimension describes how tasks are generated
and when the predictor is fitted. A simulator specifies a distribution over
graphs, mechanisms, noise, and dataset sizes, together with a label rule.
A simulator that varies few factors can simplify training, but the predictor
may perform poorly on other mechanisms or graph structures. Broader simulators
vary more factors. Training with background knowledge can also vary its
reliability and density. These choices increase training diversity, although
accuracy on unseen simulator families still needs to be evaluated.

\begin{figure*}[t]
\centering
\includegraphics[width=\textwidth]{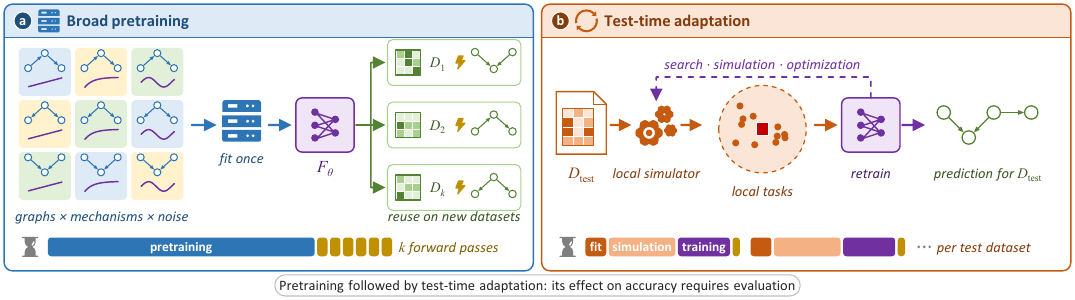}
\caption{Training and deployment costs for broad pretraining and test-time adaptation. (a) One predictor is trained on diverse simulated tasks and reused across datasets. (b) A local simulator is fitted to a test dataset, then generates local tasks for retraining the predictor. Test-time adaptation adds simulation and training costs for each test dataset. Broad pretraining can also be followed by test-time adaptation; its effect on prediction accuracy requires evaluation. Cost bars are schematic.}
\Description{A grid represents tasks with different graphs, mechanisms, and noise distributions. New datasets reuse the pretrained predictor. Panel (b) shows fitting a local simulator, generating local tasks, and retraining the predictor for one test dataset. Cost bars contrast one pretraining run with repeated simulation and training for each test dataset.}
\label{fig:broad-vs-testtime}
\end{figure*}

\parh{Training schedules.}~Training can consist of pretraining before
deployment, training for an
individual test dataset, and pretraining followed by adaptation. The first
fits a reusable model once. The second constructs a local simulator or
augmented training set from the available test data, as in ML4S, TTT-SCL, and
TICL~\cite{ma2022ml4s,deng2026ttt,chen2026ticl}. The third combines the two stages.
\F~\ref{fig:broad-vs-testtime} illustrates broad pretraining and test-time
adaptation. Test-time adaptation may reduce some mismatch with the test dataset, but
it adds simulation and optimization costs and can reproduce sampling noise.
Agreement between simulated and observed distributions alone cannot select
between observationally equivalent causal models.

\parh{Transfer to new tasks.}~Evaluation should vary the factors expected to
change at deployment.
These include changes in graph size and degree, mechanisms, noise distributions,
latent confounding, interventions, and side information. Compositional tests
combine familiar local patterns into unfamiliar larger graphs. Robustness to
one factor does not imply robustness to the others; recent work reports
substantial difficulties under compositional shift~\cite{deng2026ttt}.
\S~\ref{sec:simulators} returns to these questions using an explicit
distribution over simulated tasks.

\subsection{Method-Level Analysis}

The following comparison describes each method's inputs, prediction target,
model components, and training data. The equations summarize its training
objective or prediction procedure.

\parh{RCC.}~RCC~\cite{lopez2015towards} predicts the causal relation between
two variables. Given a bivariate sample, its classifier predicts $X\to Y$,
$Y\to X$, or a non-causal relation. RCC uses distributional features that can
represent the functional asymmetries used by A4 methods, such as independence
between a cause and the noise in the effect. Its directional predictions
therefore depend on these asymmetries being present in both training and test
data. A1--A3 alone do not identify the direction of a bivariate relation.

We focus here on RCC's bivariate setting; the source paper also discusses
extensions using multivariate context. Independently estimated pairwise
directions may conflict and do not by themselves distinguish direct effects
from confounding or indirect paths. The relation classifier is summarized as
\begin{equation}
    h_\theta\!\left(\phi(\widehat P_{X,Y})\right)
    \in \{X\!\to\!Y,\;Y\!\to\!X,\;\text{non-causal}\},
\end{equation}
where $\phi(\widehat P_{X,Y})$ denotes distributional features of a bivariate
sample. This differs from a conventional pairwise independence test because the
directional rule is learned from labeled cause-effect pairs rather than derived
from a fixed analytic statistic. In the T\"ubingen experiment, RCC trains on
synthetic pairs and tunes their generator against unlabeled test inputs;
the direction labels are used for evaluation. A separate experiment trains
on labeled ChaLearn pairs, including independent and confounded cases.

\parh{ML4S.}~ML4S~\cite{ma2022ml4s} learns the causal skeleton using a
cascade of adjacency classifiers. Under A1--A3, adjacency is identified by
the absence of a separating set. The classifiers use local CI-test statistics
and progressively prune candidate adjacencies.

Training is specific to the input dataset. A proxy discovery algorithm fits
a pseudo Bayesian network, whose conditional probability tables are estimated
from the data. Mutating this network and adjusting its parameters produces
vicinal graphs from which labeled datasets are simulated. The resulting
classifiers are then applied to the input dataset. Deployment therefore
includes proxy discovery, parameter estimation, simulation, feature extraction,
and classifier training. An adjacency decision is summarized by
\begin{equation}
    \widehat A_{ij}
    =\mathbb I\!\left[f_\theta\!\left(
      \psi(\widehat P_{\bm X},i,j)\right)>\tau\right],
\end{equation}
where $\psi$ collects the local CI evidence. A skeleton provides adjacencies;
additional orientation information is needed to recover a CPDAG.

\parh{SPOT.}~SPOT~\cite{ma2024scalable} combines a learned skeleton
posterior with per-dataset stochastic graph optimization. Its supervised
component estimates adjacency probabilities from CI statistics. These
probabilities guide a second stage of differentiable MAG learning in the
presence of latent confounding. The complete method therefore includes both
amortized prediction and graph fitting. Its learned skeleton concerns the
observed MAG, whose adjacencies need not be edges in the full latent-variable
DAG. The Markov and faithfulness conditions identify the MAG skeleton, but
a particular optimized MAG is not necessarily uniquely identified within its
PAG equivalence class.

\parh{ML4C.}~ML4C~\cite{dai2023ml4c} predicts whether an unshielded triple
(UT) is a v-structure, taking both a dataset and a supplied skeleton
$S_{\mathrm{in}}$ as inputs. The skeleton determines the triples and their
vicinities. The method's ``latent vicinity'' refers to this neighborhood
information, not to unobserved confounders. Under A1--A3, unshielded
colliders are identifiable from CI relations. Thus, this target identifies
some edge directions that cannot be recovered from a bivariate relation alone.

The triple predictions are combined using conflict resolution and graph
orientation rules. CPDAG recovery is conditional on the supplied skeleton;
a complete pipeline must also account for skeleton estimation and its errors.
Extending the procedure to latent confounding would require different
graphical assumptions and MAG/PAG orientation rules. The classifier is written as
\begin{equation}
    \widehat V_{i k j}
    =
    \mathbb{I}\!\left[
    g_\theta\!\left(\psi(\widehat P_{\bm{X}},S_{\mathrm{in}},i,k,j,\mathcal V_{ikj})\right)>\tau
    \right],
\end{equation}
where $\widehat V_{ikj}=1$ denotes the decision that an unshielded triple
$X_i-X_k-X_j$ is oriented as $X_i\to X_k\leftarrow X_j$, and
$\mathcal V_{ikj}$ denotes its latent vicinity.

\parh{DAG-EQ.}~DAG-EQ~\cite{li2020supervised} trains on simulated datasets
with known DAGs and predicts edge probabilities from the Pearson correlation
matrix of a new dataset. It then adds edges in descending probability order,
retaining only probabilities above $0.5$ and skipping any edge that would
create a cycle. This graph-assembly step is part of deployment.

Its training labels specify every edge direction. Under A1--A3 alone,
observational CI relations do not identify all of these directions.
The reported Gaussian generator uses independent unit-variance errors,
an identifying restriction on the raw linear SEM; non-Gaussian experiments
are also reported. However, the correlation encoder can discard information
that identifies direction. For example, opposite two-node models
$X=N_X,\ Y=2X+N_Y$ and $Y=N_Y,\ X=2Y+N_X$, with independent standard
Gaussian errors, have different covariance matrices but the same correlation
matrix. Identifiability from the full distribution therefore does not ensure
identifiability from this encoder's input. The training objective is summarized by
\begin{equation}
    \min_\theta
    \sum_m\sum_{i\ne j}
    \operatorname{CE}\!\left(A^{(m)}_{ij},
    f_\theta(\widehat{\operatorname{Corr}}^{(m)}_{\bm{X}})_{ij}\right),
\end{equation}
where $A^{(m)}$ is the DAG label generated by the simulator for task $m$.
This objective fits the edge predictor across tasks; at test time, the fitted
predictor supplies edge scores to the cycle-avoiding assembly rule.

\parh{CSIvA.}~CSIvA~\cite{ke2022learning} learns from DAG labels using an
autoregressive graph decoder. Each graph decision is conditioned on earlier
decisions, so the decoder can represent dependencies among edges. The
published decoder does not enforce acyclicity, so sequential generation
alone does not guarantee a valid DAG.

Sequential decoding does not determine whether an edge direction is
identifiable. The predicted directions may depend on functional restrictions
or other orientation information in the training tasks, but may also reflect
an ordering convention used by the simulator. The decoder factorizes the
conditional graph distribution as
\begin{equation}
    p_\theta(G\mid \widehat P_{\bm{X}})
    =
    \prod_{t=1}^{T}
    p_\theta(e_t \mid \widehat P_{\bm{X}}, e_{<t}),
\end{equation}
where $e_t$ denotes the $t$-th graph-construction decision. Unlike independent
edge classification, this factorization lets later edge decisions condition on
the partially generated graph.

\parh{AVICI.}~AVICI~\cite{lorch2022amortized} estimates posterior edge
probabilities with a transformer trained on diverse simulated tasks. The
trained predictor can be applied to a new dataset without further training.
Its variational outputs approximate posterior edge marginals under the
simulator prior, rather than a flexible joint posterior over graphs.
AVICI permits general directed causal structures. Its LINEAR and RFF
experiments impose an acyclicity penalty, whereas its gene-regulatory
experiments use steady-state data from a stochastic dynamical simulator.
The penalty can affect the fitted approximation and does not guarantee
acyclicity under approximate optimization.

When the data do not distinguish members of an equivalence class, posterior
edge probabilities also depend on their prior probabilities. Functional
restrictions such as A4 may provide additional information that identifies
directions. Calibrated edge marginals can help assess uncertainty about
individual edges. In an acyclic SCM setting, obtaining a valid DAG or CPDAG
requires an appropriate graph-construction procedure; PAG inference would
additionally require a model and labels for latent confounding. The marginal
approximation is summarized as
\begin{equation}
    q_\theta(A_{ij}=1 \mid \widehat P_{\bm{X}})
    \approx
    p(A_{ij}=1 \mid \widehat P_{\bm{X}}, \Pi_{\mathrm{sim}}),
\end{equation}
where $\Pi_{\mathrm{sim}}$ denotes the simulator prior over graphs and
mechanisms. The model estimates an edge marginal under this prior; a
thresholded collection of marginals still needs a graph construction rule.

\parh{BCNP.}~BCNP~\cite{dhirmeta} learns an approximation
$q_\phi(G\mid D)$ to the graph posterior of a Bayesian causal model. Its
decoder samples a permutation matrix and a triangular Bernoulli edge matrix.
The permutation specifies a topological ordering, and the triangular matrix
selects edges consistent with that ordering. Each resulting sample is
therefore a DAG, allowing BCNP to represent uncertainty over complete graphs.

The posterior approximation depends on the Bayesian causal model used to
generate training tasks. Evaluation on a new task should therefore assess
whether the predicted distribution remains calibrated when the graph,
mechanism, or noise assumptions change, including changes in Markov,
faithfulness, or causal sufficiency assumptions. The decoder samples graphs as
\begin{equation}
    \begin{aligned}
    \widetilde Q_s &\sim \operatorname{GS}_{\lambda}\!\left(\Theta_\phi(\widehat P_{\bm{X}})\right),\\
    Q_s &=\operatorname{Hungarian}(\widetilde Q_s),\\
    A_s &\sim \operatorname{Bern}\!\left(\Phi_\phi(\widehat P_{\bm{X}})\right),
    \qquad [\Phi_\phi]_{ij}=0\ \text{for } i\ge j,\\
    G_s &= Q_s A_s Q_s^\top .
    \end{aligned}
\end{equation}
Here $\widetilde Q_s$ is a soft Gumbel-Sinkhorn draw at temperature
$\lambda>0$, and Hungarian hardening produces a discrete permutation $Q_s$
for the forward pass. A straight-through estimator supplies gradients.
Conjugating the triangular binary matrix $A_s$ by this discrete permutation
guarantees a DAG. Mixing over the shared sampled order induces dependencies
among graph edges; shared deterministic network parameters alone would not
make independent Bernoulli draws dependent.

\parh{SiCL.}~SiCL~\cite{zhanglearning} predicts a CPDAG by separately
estimating the skeleton and v-structures. These labels correspond to the
graph features identifiable from observational CI relations under A1--A3.
Conflict-resolution heuristics and orientation rules combine the local
adjacency and collider predictions into a CPDAG estimate.

Evaluation should score this CPDAG target. Scoring a particular DAG
representative would also reward or penalize directions that the assumed
observational model does not identify. The prediction procedure is written as
\begin{equation}
    \widehat{\mathcal C}
    =
    \operatorname{CPDAG}\!\left(
    \widehat S_\theta(\widehat P_{\bm{X}}),
    \widehat V_\theta(\widehat P_{\bm{X}})
    \right),
\end{equation}
where $\widehat S_\theta$ predicts the skeleton and $\widehat V_\theta$ predicts
v-structures. The decoder combines these estimates into a CPDAG.

\parh{Kode.}~Kode~\cite{xu2026knowledge} takes observational data and
background knowledge as inputs. Its knowledge matrix represents whether a
causal influence is possible, impossible, or unknown. A possible influence
can indicate an indirect ancestral relation, so the matrix does not directly
specify edge labels. An encoder processes the data and knowledge matrix, and
the encoded knowledge adds a weak bias before graph decoding. A decoder based
on BCNP then samples DAGs using permutation and triangular edge matrices.

The knowledge input provides information about orientation and reachability
in addition to the observational data. However, incorrect knowledge or too
many asserted relations can bias the predictions. Kode varies the strength
and sparsity of the knowledge input during pretraining. Varying its coverage
and strength does not by itself establish robustness to arbitrary false
knowledge. The predictor can be written as
\begin{equation}
    \begin{aligned}
    K_{ij} &\in \{-1,0,1\}
    \quad
    (\text{impossible, unknown, may hold}),\\
    H_\theta &= E_\theta(\widehat P_{\bm{X}},K),
    \qquad
    \widetilde H_\theta = H_\theta+\alpha\,R_\theta(K),\\
    \widetilde Q_s &\sim \operatorname{GS}_{\lambda}\!\left(\Theta_\theta(\widetilde H_\theta)\right),\\
    Q_s &=\operatorname{Hungarian}(\widetilde Q_s),\qquad
    L_s \sim \operatorname{Bern}\!\left(\Phi_\theta(\widetilde H_\theta)\right),
    \\
    [\Phi_\theta]_{ij}&=0\ \text{for } i\le j,\qquad
    G_s = Q_s L_s Q_s^\top .
    \end{aligned}
\end{equation}
Here $K$ is the encoded reachability prior, $H_\theta$ is the dual-source
representation, $\alpha$ controls the strength of the knowledge bias, $Q_s$ is
a hardened permutation matrix, and $L_s$ is a sampled lower-triangular edge
matrix. As in BCNP, hardening preserves discreteness in the forward pass
while a straight-through estimator supplies gradients. The knowledge bias
influences graph sampling while keeping
reachability information distinct from direct-edge labels.

\parh{SEA.}~SEA~\cite{wu2025sample} combines estimates from classical
discovery algorithms with a learned graph predictor. Its
sample-estimate-aggregate procedure samples small subsets of variables and
data batches, then runs a classical discovery algorithm on each subset. An
aggregator with axial attention combines these marginal graph estimates with
global statistics, such as inverse covariance, to predict the complete graph.
Training uses the complete synthetic graph as the label. The reported
implementations use FCI for observational data and GIES for interventional
data; the architecture also allows other sampling procedures, marginal
estimators, and global statistics.

The marginal graph estimates must be interpreted under the assumptions of
the classical estimator. FCI can represent latent confounding caused by
omitting variables from a subset. GIES uses interventional data and depends
on its score assumptions. The aggregator learns to combine these estimates,
so it can reuse the classical computations. However, errors in the marginal
estimates, inadequate coverage of variable subsets, or differences between
the marginal and final graph targets can affect its predictions. The
procedure is summarized by
\begin{equation}
    \begin{aligned}
    \rho &= s(D_0),\\
    E'_t &= f\!\left(D_t[S_t]\right),\qquad t=1,\ldots,T,\\
    \widehat E
    &= a_\theta\!\left(\rho,\{(S_t,E'_t)\}_{t=1}^{T}\right),
    \end{aligned}
\end{equation}
where $s$ computes global statistics on a batch $D_0$, $S_t$ is a sampled node
subset, $D_t[S_t]$ is the restricted batch, $f$ is the marginal causal discovery
algorithm, and $E'_t$ is the estimated marginal graph over $S_t$. The
aggregator conditions jointly on the global statistics and indexed
marginal estimates. Its deployment cost includes the classical estimators run
on those subsets.

\parh{ADAG.}~ADAG~\cite{yin2025learning} trains a predictor with an
attention-based kernel across several domains defined by structural equation
models (SEMs). It predicts a weighted adjacency matrix that represents both
the graph structure and linear SEM coefficients. Training minimizes a SEM
reconstruction loss subject to an acyclicity constraint, without using
ground-truth DAG labels.

ADAG considers two training settings. In heterogeneous domains, the graph is
shared while mechanisms vary. In order-consistent domains, the DAGs differ
but share a topological ordering. The trained predictor uses these shared
structural properties and can be applied to a new domain without retraining.
The paper mainly studies linear SEMs under these assumptions. Identifying
edge directions still depends on the noise assumptions and the relations
between domains; linear additivity or acyclicity alone is insufficient. The
training objective is
\begin{equation}
    \begin{gathered}
    \min_{\theta}\sum_{m=1}^{M}
    \mathcal L_{\mathrm{SEM}}\!\left(X^{(m)},K_\theta(X^{(m)})\right),\\
    \text{subject to}\quad h\!\left(K_\theta(X^{(m)})\right)=0,
    \quad m=1,\ldots,M.
    \end{gathered}
\end{equation}
Here $K_\theta$ is the attention-based nonlinear kernel map from domain data to
weighted adjacency, $\mathcal L_{\mathrm{SEM}}$ is the linear-SEM
reconstruction objective with sparsity regularization, and $h(\cdot)$ is a
continuous DAG constraint. The fitted map $K_\theta$ is reused for
new domains through a forward pass.

\parh{Arrow.}~Arrow~\cite{thompson2026arrow} predicts an undirected skeleton
and a topological order, then orients the skeleton edges according to that
order. This construction guarantees an acyclic output. A pretrained
transformer computes the skeleton and ordering scores for a new observational
dataset without further training. The training simulator varies graph
families, mechanisms, noise distributions, and dataset sizes. Supervision uses
the directed adjacency matrix through a composite likelihood of directed
edges. Skeleton probabilities and order scores are decoder outputs; a single
chosen topological order is not the training label.

Although the decoder guarantees a DAG, the interpretation of its directions
depends on the mechanisms and priors used in training and whether they apply
to the test dataset. The decoder constructs the adjacency matrix as
\begin{equation}
    \widehat A_{ij}
    =
    \widehat S_{ij}\,
    \mathbb{I}\!\left[\widehat\pi_i < \widehat\pi_j\right],
    \qquad
    (\widehat S,\widehat\pi)
    =
    F_\theta(\widehat P_{\bm{X}}),
\end{equation}
where $\widehat S$ is the predicted skeleton and $\widehat\pi_i$ is the rank
of node $i$ in the selected total order. Orienting each skeleton edge according to this order
guarantees acyclicity. Identifying the order from data requires additional
orientation information.

\parh{TTT-SCL.}~TTT-SCL~\cite{deng2026ttt} trains an SCD predictor
for an individual test dataset. The study reports that pretrained predictors
can lose accuracy under distribution shift, especially when familiar local
patterns occur in unfamiliar graph structures. It also reports differences
in accuracy between synthetic benchmarks and real or pseudo-real data.
TTT-SCL addresses this problem by generating training tasks matched to the
test dataset. Its graph-generation procedure scores candidate graphs using an
alignment-of-distribution criterion and a sparsity term. Stochastic graph
search refines these graphs so that simulated datasets resemble the test
dataset. The resulting local simulator generates tasks for retraining an
existing SCD predictor.

TTT-SCL changes the training regime while retaining the base predictor's
target, encoder, and decoder. Because the local simulator is fitted to a
single dataset, it can also fit sampling noise. Moreover, matching the
observed distribution cannot resolve directions that are not identifiable
without functional restrictions or external information.
\S~\ref{sec:simulators} examines these limits in terms of the local
task distribution.

\parh{TICL.}~TICL~\cite{chen2026ticl} uses test-time training when
interventional data are available but the intervention targets are unknown.
It combines observational and interventional data through the joint causal
inference (JCI) protocol. A self-augmentation procedure generates training
data for the test dataset. A supervised procedure based on PC then predicts
the skeleton and orients edges using JCI priors and Meek's rules.

TICL targets an equivalence class under assumptions based on A1--A3, with
additional orientations supported by interventional data and JCI assumptions.
These data can identify directions left unresolved by observational CI
relations. Recovery therefore depends on which interventions are available
and whether the intervention assumptions hold.

\subsection{Synthesis and Evaluation Implications}

Evaluation metrics should match each method's prediction target.
Skeleton accuracy measures adjacency recovery, CPDAG accuracy includes
identifiable orientations, and full-DAG accuracy also scores directions that
may depend on functional restrictions or the simulator prior. Posterior
methods additionally require uncertainty assessment. A common graph metric
can obscure these distinctions if the graph target is left unspecified.

To compare methods, a study should state the prediction target and the
training and test assumptions, then explain how model outputs are converted
into the graph being scored. These details help distinguish recovery of
identifiable graph features from agreement with directions favored by a
particular simulator.

Training distributions can differ even under the same causal assumptions. Two
simulators can both generate Markov, faithful, causally sufficient tasks
while using different mechanisms, graph priors, and sample sizes. They can
therefore induce different prediction rules and different failure modes.
The next section examines these differences in more detail.

\section{Simulators as Sources of Supervision}
\label{sec:simulators}

Simulators generate the datasets and structural labels used in the workflow
of \S~\ref{sec:scl-workflow}. This section examines how their assumptions
affect what a predictor can learn. We first define the distribution over
training tasks, then discuss which graph features are identifiable within
that distribution and what can change at deployment.

\subsection{Simulators as Data-and-Label Generators}

Let $\eta$ denote a complete data-generating specification sampled from a
simulator prior $\Pi_{\mathrm{sim}}$:
\begin{equation}
    \eta
    =
    (G, \bm f, P_{\bm N}, \mathcal E, \bm H, K, n)
    \sim \Pi_{\mathrm{sim}},
    \qquad
    D \sim P_{\eta}^{(n)} .
\end{equation}
Here $G$ is the causal graph, $\bm f$ denotes structural mechanisms,
$P_{\bm N}$ denotes the exogenous noise distribution, $\mathcal E$ denotes available
interventions or environments, $\bm H$ denotes latent variables, $K$ denotes
optional background knowledge, and $n$ is the sample size. The prior
$\Pi_{\mathrm{sim}}$ induces a joint distribution over datasets and structural
labels. If $\tau(\eta)$ is the target provided by the
simulator, such as a DAG, CPDAG, PAG, skeleton, or edge indicator, the empirical
training objective approximates
\begin{equation}
    \theta^\star
    \in
    \arg\min_{\theta}
    \mathbb E_{\eta\sim\Pi_{\mathrm{sim}}}
    \mathbb E_{D\sim P_\eta^{(n)}}
    \left[
        \ell\!\left(F_\theta(D),\tau(\eta)\right)
    \right].
    \label{eq:sim-training-objective}
\end{equation}
With an appropriate probabilistic loss, these paired datasets and labels can
train a predictor of the conditional graph distribution or its marginals.
Each sampled graph provides a training label, while the posterior describes
uncertainty over graphs given a dataset and the simulator prior. When side
information $K$ is used, the predictor and conditional expectations below also
condition on $K$.

\begin{figure*}[t]
\centering
\includegraphics[width=\textwidth]{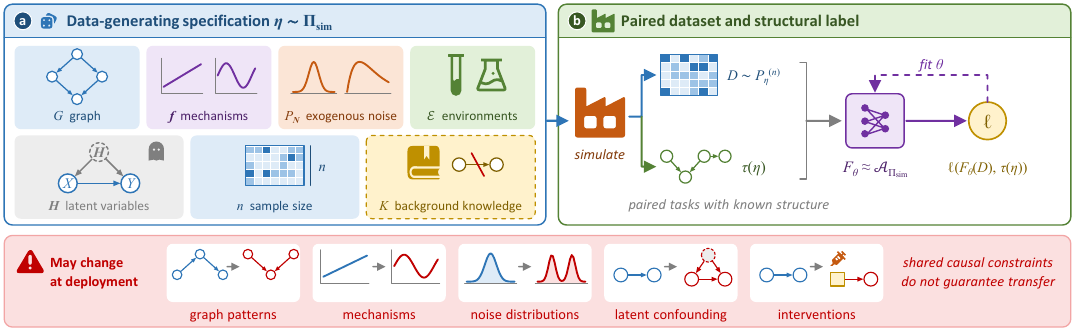}
\caption{Generation of training datasets and structural labels. (a) A data-generating specification $\eta$ includes a graph, mechanisms, exogenous noise, environments, latent variables, sample size, and optional background knowledge. (b) The simulator generates a dataset $D$ and structural label $\tau(\eta)$ from this specification to train $F_\theta$. The bottom row illustrates factors that may change at deployment. Sharing causal constraints between training and test distributions alone does not guarantee transfer.}
\Description{Seven tiles show the components of a data-generating specification. A simulator turns it into a paired dataset and structural label used to fit a predictor. A bottom strip shows changes in graph patterns, mechanisms, noise distributions, latent confounding, and interventions at deployment.}
\label{fig:simulator-object}
\end{figure*}

The learned predictor therefore approximates the prediction rule defined by
the simulator prior and the loss (\F~\ref{fig:simulator-object}(b)):
\begin{equation}
\begin{aligned}
    F_\theta(D)
    &\approx
    \mathcal A_{\Pi_{\mathrm{sim}}}(D),
    \\
    \mathcal A_{\Pi_{\mathrm{sim}}}(D)
    &=
    \arg\min_{t}
    \mathbb E\!\left[\ell(t,\tau(\eta))\mid D,\Pi_{\mathrm{sim}}\right].
\end{aligned}
    \label{eq:sim-conditioned-rule}
\end{equation}
A supervised predictor can use distributional features that are predictive
of its labels under this joint distribution. Classical functional-model and Bayesian
methods can also use information beyond CI relations; the supervised approach
learns a reusable prediction rule from sampled tasks.

\parh{Simulators used by individual methods.}~\T~\ref{tab:simulator-instantiations} compares
the data-generating distributions and structural label rules that define the
training tasks of representative discovery methods.

\begin{table*}[!t]
\centering
\caption{Simulators used by representative supervised
causal discovery methods. Here $\eta$ denotes one simulated data-generating
instance; $\tau(\eta)$ denotes the structural target. For ADAG, the
structure is an evaluation target, not a supervised training label.}
\label{tab:simulator-instantiations}
\setlength{\jot}{0pt}
\setlength{\tabcolsep}{2pt}
\renewcommand{\arraystretch}{1.12}
\setlength{\extrarowheight}{3pt}
\newcommand{\simtwomethod}[1]{#1}
\newcommand{\simthreemethod}[1]{#1}
\newcommand{\dagtarget}{\tau(\eta)=A(G)}
\scriptsize
\begin{tabularx}{\textwidth}{@{}
    >{\footnotesize\raggedright\arraybackslash}p{0.13\textwidth}
    >{\footnotesize\raggedright\arraybackslash}p{0.23\textwidth}
    >{\scriptsize\arraybackslash}X@{}}
\toprule
{\footnotesize\textbf{Method}} & {\footnotesize\textbf{Simulation Target}} & {\footnotesize\textbf{Instantiation}} \\
\midrule
\simtwomethod{RCC~\cite{lopez2015towards}} & Synthetic Bivariate ANM &
\(\begin{gathered}
\eta=(P_X,f,P_N,n)\sim\Pi_{\mathrm{ANM}},\quad
X\sim\sum_k\pi_k\mathcal N(\mu_k,\sigma_k^2),\quad Y=f(X)+N,\\
N\ci X,\quad N\sim\mathcal N(0,\sigma_N^2),\quad
f\sim\Pi_{\mathrm{spline}},\quad
\tau(\eta)=X\!\to\!Y,\quad (Y,X)\mapsto Y\!\to\!X .
\end{gathered}\) \\[4pt]
\addlinespace[3pt]
& Transductive T\"ubingen Setting &
Generator parameters are fitted against unlabeled test-pair embeddings.
Training labels come from synthetic pairs; T\"ubingen directions are
evaluation labels. A separate ChaLearn experiment uses labeled pairs. \\[4pt]
\midrule
ML4S~\cite{ma2022ml4s} & Dataset-Specific Discrete Skeleton &
\(\begin{gathered}
(G_0,\theta_0)=\operatorname{FitBN}(D_{\mathrm{test}}),\quad
G'\sim q_{\mathrm{vic}}(\cdot\mid G_0),\\
\theta'=\operatorname{AdjustCPT}(G',G_0,\theta_0),\quad
D'\sim P_{G',\theta'}^{(n)},\\
\eta=(G',\theta',n),\qquad
\tau(\eta)=\operatorname{skel}(G').
\end{gathered}\) \\[4pt]
\midrule
SPOT~\cite{ma2024scalable} & Linear Gaussian MAG Skeleton &
\(\begin{gathered}
\eta=(G,H,\theta,n)\sim\Pi_{\mathrm{ADMG}},\quad
M=\operatorname{MAG}(G,H),\\
D\sim P_{\eta}^{(n)},\quad
\tau(\eta)=\operatorname{skel}(M).
\end{gathered}\) \\[4pt]
\midrule

ML4C~\cite{dai2023ml4c} & Discrete V-Structure &
\(\begin{gathered}
\eta=(G,\theta,n)\sim\Pi_{\mathrm{ML4C}},\quad
G\sim\Pi_{\mathrm{ER/SF}},\quad
D\sim\prod_i P_{\theta_i}(X_i\mid X_{\operatorname{Pa}_G(i)}),\\
(i,k,j)\in\operatorname{UT}(G),\quad
\tau(\eta)=\operatorname{vstr}(G).
\end{gathered}\) \\[4pt]
\midrule
DAG-EQ~\cite{li2020supervised} & Equal-Variance Gaussian DAG &
\(\begin{gathered}
\eta=(G,B_G,n)\sim\Pi_{\mathrm{LG}},\quad
G\sim\Pi_{\mathrm{ER/SF}},\quad
X=B_G^\top X+\varepsilon,\quad
\varepsilon\sim\mathcal N(0,I),\\
D=\{x^{(r)}\}_{r=1}^n,\qquad
\tau(\eta)=A(G).
\end{gathered}\) \\[4pt]
\midrule
\simthreemethod{CSIvA~\cite{ke2022learning}} & Dirichlet Categorical DAG &
\(\begin{gathered}
\eta=(G,\theta,\mathcal E,n)\sim\Pi_{\mathrm{cat}},\quad
G\sim\Pi_{\mathrm{ER}},\quad
\theta_{j\mid\mathrm{pa}}\sim\operatorname{Dir}(\alpha),\quad
X_j\mid X_{\operatorname{Pa}_G(j)}\sim\operatorname{Cat}(\theta_{j\mid\mathrm{pa}}),\\
D=\{D^{\mathrm{obs}},D^{\mathrm{int}}\},\quad \dagtarget .
\end{gathered}\) \\[4pt]
\addlinespace[3pt]
& Linear Continuous DAG &
\(\begin{gathered}
\eta=(G,W,P_{\bm N},\mathcal E,n)\sim\Pi_{\mathrm{lin}},\quad
G\sim\Pi_{\mathrm{ER}},\quad
X_j=\sum_{i\in\operatorname{Pa}_G(j)}w_{ij}X_i+\varepsilon_j,\\
D=\{D^{\mathrm{obs}},D^{\mathrm{int}}\},\quad \dagtarget .
\end{gathered}\) \\[4pt]
\addlinespace[3pt]
& MLP Nonlinear DAG &
\(\begin{gathered}
\eta=(G,\bm f,P_{\bm N},\mathcal E,n)\sim\Pi_{\mathrm{MLP}},\quad
G\sim\Pi_{\mathrm{ER}},\quad
X_j=f_j^{\mathrm{MLP}}(X_{\operatorname{Pa}_G(j)},\varepsilon_j),\\
D=\{D^{\mathrm{obs}},D^{\mathrm{int}}\},\quad \dagtarget .
\end{gathered}\) \\[4pt]
\midrule
\simtwomethod{AVICI~\cite{lorch2022amortized}} & Linear/RFF DAG &
\(\begin{gathered}
\eta=(G,\bm f,P_{\bm N},\mathcal E,n)\sim\Pi_{\mathrm{AVICI}},\quad
G\sim\Pi_{\mathrm{ER/SF}},\quad
X_i=f_i(X_{\operatorname{Pa}_G(i)})+\varepsilon_i,\\
f_i\in\{\beta_i^\top x,\operatorname{RFF}_i(x)\},\quad
\varepsilon_i\sim\Pi_N,\quad D^{(e)}\sim P_{\eta}^{(e)},\\
\dagtarget .
\end{gathered}\) \\[4pt]
\addlinespace[3pt]
& Gene-Regulatory Dynamics &
\(\begin{gathered}
\eta=(G,\theta,\mathcal E,n)\sim\Pi_{\mathrm{SERGIO}},\quad
G\subseteq G_{\mathrm{reg}},\quad
Z\sim\operatorname{SERGIO}(G,\theta),\\
X=h_{\mathrm{platform}}(Z)+\xi,\quad
D^{(e)}\sim P_{\eta}^{\operatorname{KO}(e)},\quad
\tau(\eta)=A(G).
\end{gathered}\) \\[4pt]
\midrule
BCNP~\cite{dhirmeta} & Bayesian Causal DAG &
\(\begin{gathered}
\eta=(G,\theta,n)\sim p(G)p(\theta\mid G)p(n),\quad
G\sim p(G),\quad \theta\sim p(\theta\mid G),\quad
D\sim p_{\theta}(X\mid G),\\
\dagtarget .
\end{gathered}\) \\[4pt]
\midrule
SiCL~\cite{zhanglearning} & ER/SF Structural Labels &
\(\begin{gathered}
\eta=(G,\bm f,P_{\bm N},n)\sim\Pi_{\mathrm{SiCL}},\quad
G\sim\Pi_{\mathrm{ER/SF}}\ (\text{continuous}),\quad G\sim\Pi_{\mathrm{SF}}\ (\text{discrete}),\\
D\sim P_{\eta},\\
f_i\in\{\beta_i^\top x,\operatorname{RFF}_i(x),\operatorname{CPT}_i\},\quad
\tau(\eta)=\bigl(\operatorname{skel}(G),\operatorname{vstr}(G)\bigr),
\quad \mathcal C=\operatorname{CPDAG}(\tau(\eta)).
\end{gathered}\) \\[4pt]
\midrule
\simtwomethod{Kode~\cite{xu2026knowledge}} & ER Synthetic DAG &
\(\begin{gathered}
\eta=(G,\bm f,P_{\bm N},n)\sim\Pi_{\mathrm{SCM}},\quad
G\sim\Pi_{\mathrm{ER}},\quad \bm f\in\{\mathrm{lin.\ het.},\ldots\},\quad
D\sim P_\eta^{(n)},\quad
\tau(\eta)=G .
\end{gathered}\) \\[4pt]
\addlinespace[3pt]
& Knowledge-Augmented DAG &
\(\begin{gathered}
\eta=(G,\bm f,P_{\bm N},K,n)\sim\Pi_{\mathrm{Kode}},\quad
K_{ij}\in\{-1,0,1\},\quad
K\sim q_{\alpha,s}(K\mid \operatorname{Anc}(G)),\\
D\sim P_\eta^{(n)},\quad \tau(\eta)=G .
\end{gathered}\) \\[4pt]
\midrule
\simtwomethod{SEA~\cite{wu2025sample}} & Observational DAG &
\(\begin{gathered}
\eta=(G,\bm f,P_{\bm N},n)\sim\Pi_{\mathrm{SEA}}^{\mathrm{obs}},\quad
G\sim\Pi_{\mathrm{ER/SF}},\quad
\bm f\in\{\mathrm{lin.},\mathrm{NN\ add.},\ldots\},\quad D\sim P_{\eta}^{\mathrm{obs}},\\
\dagtarget .
\end{gathered}\) \\[4pt]
\addlinespace[3pt]
& Interventional DAG &
\(\begin{gathered}
\eta=(G,\bm f,P_{\bm N},\mathcal E,n)\sim\Pi_{\mathrm{SEA}}^{\mathrm{int}},\quad
G\sim\Pi_{\mathrm{ER/SF}},\quad
D=\{D^{\mathrm{obs}},D^{\operatorname{do}(j)}\}_{j\in V},\\ \dagtarget .
\end{gathered}\) \\[4pt]
\midrule
\simtwomethod{ADAG~\cite{yin2025learning}} & Shared-Graph Linear DAG
&
\(\begin{gathered}
\eta=(A,\{W_m,P_m,n_m\}_{m=1}^M)\sim\Pi_{\mathrm{ADAG}},\quad A_m=A,\quad B_m=A\odot W_m,\\
X^{(m)}=B_m^\top X^{(m)}+\varepsilon^{(m)},\quad \tau(\eta)=A .
\end{gathered}\) \\[4pt]
\addlinespace[3pt]
& Order-Consistent Linear DAG &
\(\begin{gathered}
\eta=(\pi,\{A_m,W_m,P_m,n_m\}_{m=1}^M)\sim\Pi_{\mathrm{ADAG}},\quad\pi_m=\pi,\quad A_m\sim\Pi(A\mid \pi),\\
B_m=A_m\odot W_m,\quad
X^{(m)}=B_m^\top X^{(m)}+\varepsilon^{(m)},\quad
\tau(\eta)=\{A_m\}_{m=1}^M.
\end{gathered}\) \\[4pt]
\midrule
Arrow~\cite{thompson2026arrow} & ER/SF Synthetic DAG &
\(\begin{gathered}
\eta=(G,\bm f,P_{\bm N},n,d)\sim\Pi_{\mathrm{Arrow}},\quad
G\sim\Pi_{\mathrm{ER/SF}},\quad \bm f\in\{\mathrm{lin.},\mathrm{MLP}\},\\
P_{\bm N}\in\{\mathcal N,\mathcal U,\operatorname{Beta}\},\quad
D\sim P_\eta^{(n)},\qquad
\tau(\eta)=A(G).
\end{gathered}\) \\[4pt]
\bottomrule
\end{tabularx}
\end{table*}

\T~\ref{tab:simulator-instantiations} lists selected training settings,
not every experiment. DAG-EQ also reports exponential, Gumbel, and Poisson
noise; CSIvA additionally studies other continuous and discrete mechanisms.
SiCL tests its continuous ER/SF-trained models on WS/SBM graphs and its
discrete SF-trained model on ER graphs. These test families are distinct from
the training generators shown in the table.

The simulators in \T~\ref{tab:simulator-instantiations} vary in their graph
distributions, mechanisms, and available data. Graphs range from bivariate
cause-effect pairs to larger DAGs, latent-variable graphs, and biological
regulatory networks. Mechanisms include linear Gaussian, categorical, neural,
and random-feature models, as well as domain-specific simulators such as
SERGIO~\cite{dibaeinia2020sergio}. Training data can include observations,
interventions, several domains, or labeled pairs from existing benchmarks.
Unlabeled benchmark inputs can also guide simulation, as in RCC.

The structural label also varies: it can be a direction, skeleton, set of
v-structures, CPDAG, or adjacency matrix. Arrow's skeleton/order factorization
specifies its decoder, while its supervised label is the directed graph.
In the table, $\operatorname{MAG}(G,H)$ denotes a MAG representing the observed
CI model after omitting $H$; it need not equal the ADMG latent projection.
ADAG uses the displayed graph targets for evaluation and trains its predictor
by reconstruction, as described in \S~\ref{sec:implications-learning}.

\subsection{Simulator-Induced Constraints and the Benefit of Supervision}

\begin{figure*}[t]
\centering
\includegraphics[width=\textwidth]{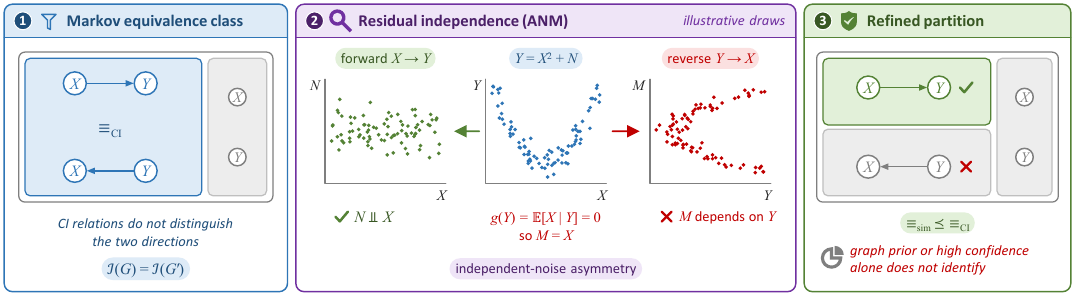}
\caption{Functional restrictions can distinguish Markov-equivalent directions. The illustrative draws use $X\sim\mathcal U[-2,2]$ and $Y=X^2+N$, with independent $N\sim\mathcal N(0,0.4^2)$. The forward residual is $N$. Symmetry gives the reverse population regression $g(Y)=\mathbb E[X\mid Y]=0$, whose residual $M=X$ depends on $Y$. Identification requires the functional-model conditions; a graph prior or high confidence alone does not supply them.}
\Description{Two opposite arrows share a Markov equivalence class. Three scatter plots of illustrative draws show the observations, the forward residual, and the reverse residual. A refined partition separates the two directions, while a graph prior or high confidence alone does not identify the direction.}
\label{fig:mec-refinement}
\end{figure*}

\parh{From Markov equivalence to simulator-induced constraints.}~Under A1--A3,
the observational information used by classical constraint-based discovery is
largely summarized by the conditional-independence model
\begin{equation}
    \mathcal I(P_{\bm{X}})
    =
    \{(A,B,C): X_A \ci X_B \mid X_C \}.
\end{equation}
Two DAGs are Markov equivalent when they imply the same such set, or
equivalently when they have the same skeleton and unshielded colliders. This
equivalence relation can be written as
\begin{equation}
    G \equiv_{\mathrm{CI}} G'
    \quad\Longleftrightarrow\quad
    \mathcal I(G)=\mathcal I(G').
\end{equation}
The simulator can add information beyond this CI model, but only through the
regularities actually built into its data-generating family. Let
$\mathcal C_{\mathrm{sim}}$ denote the collection of distributional regularities
that hold across the simulator family:
\begin{equation}
    \mathcal C_{\mathrm{sim}}
    =
    \{c_\ell:\ c_\ell(P_\eta)=0
      \text{ for all } \eta\in\operatorname{supp}(\Pi_{\mathrm{sim}})\}.
\end{equation}
Constraints shared across all sampled graphs can be too weak to describe
the directional information available in a task. For a fixed graph $G$, let
$\mathcal C_{\mathrm{sim}}(G)$ denote constraints that hold for every model
with that graph in the simulator's support. We can compare graphs using the
pair $(\mathcal I(G),\mathcal C_{\mathrm{sim}}(G))$:
\begin{equation}
\begin{aligned}
 G\equiv_{\mathrm{sim}}G'\quad\Longleftrightarrow\quad
 &\mathcal I(G)=\mathcal I(G'),\\
 &\mathcal C_{\mathrm{sim}}(G)=\mathcal C_{\mathrm{sim}}(G').
\end{aligned}
\end{equation}
This definition refines the CI partition whenever the additional constraints
differ within a Markov equivalence class
($\equiv_{\mathrm{sim}}\preceq\equiv_{\mathrm{CI}}$;
\F~\ref{fig:mec-refinement}). Different constraint sets need not imply
non-overlapping distribution families: exceptional distributions may satisfy
both. Identifiability still requires the implication in
\eqref{eq:target-identifiability}, restricted to the simulator's model class.
A graph prior can also favor one graph without making it identifiable.

\begin{figure*}[t]
\centering
\includegraphics[width=\textwidth]{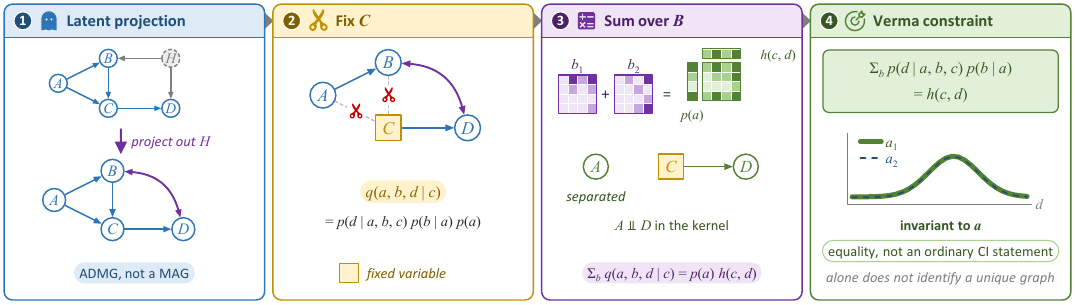}
\caption{The Verma constraint in four steps. Projecting out a common cause of $B$ and $D$ gives the bidirected edge in the ADMG. Fixing $C$ removes its incoming edges and yields a derived kernel. Summing this kernel over $B$ separates $A$ and $D$ conditional on fixed $C$; equivalently, the displayed weighted sum is invariant to $a$. The squares denote fixed variables; heat maps are schematic. This constraint alone does not establish a uniquely identified graph.}
\Description{A latent common cause is projected into an ADMG. Fixing C removes incoming edges, kernel slices summed over B factorize into p(a) times h(c,d), and a plot shows that the weighted sum is invariant to a.}
\label{fig:verma-running-example}
\end{figure*}

Several familiar examples fit this view. Functional-model simulators impose
asymmetries such as additive independent noise or non-Gaussian linear noise,
which can distinguish directions that are Markov equivalent under A1--A3.
Multi-environment simulators impose invariance constraints, for example
\begin{equation}
\begin{aligned}
    P^{(e)}(X_j\mid \operatorname{Pa}_G(X_j))
    &=
    P^{(e')}(X_j\mid \operatorname{Pa}_G(X_j)),
    \\
    &\text{for environments } e,e' \notin \mathcal E_j,
\end{aligned}
\end{equation}
while allowing mechanisms targeted by interventions to vary. Latent-variable
SCM simulators can also imply equality constraints on the observed margin that
are not conditional independencies. As a running example, consider latent-variable SCMs whose latent projection
over $(A,B,C,D)$ is the Verma acyclic directed mixed graph (ADMG), with
$A\to B$, $A\to C$, $B\to C$, $C\to D$, and $B\leftrightarrow D$.
This ADMG is not a MAG: $B$ is an ancestor of $D$ as well as being connected to
it by a bidirected edge. The distinction matters because an ADMG retains
information used by the nested Markov model that need not be expressed by
ordinary observed CI relations~\cite{shpitser2008dormant,richardson2023nested}.
\F~\ref{fig:verma-running-example} shows the graph and the kernel obtained by
fixing $C$.
For a positive observed distribution, the dormant-independence argument
fixes $C$ to obtain the kernel
\begin{equation}
    q_\eta(a,b,d\mid c)
    =
    p_\eta(d\mid a,b,c)\,p_\eta(b\mid a)\,p_\eta(a),
\end{equation}
and, after marginalizing $B$, the fixed graph implies $A \ci D$ in this kernel.
Therefore every sampled $\eta\in\operatorname{supp}(\Pi_{\mathrm{sim}})$
satisfies
\begin{equation}
    \sum_b
    p_\eta(d\mid a,b,c)\,p_\eta(b\mid a)
    =
    h_\eta(c,d),
    \label{eq:verma-running-example}
\end{equation}
so the left-hand side is invariant to $a$. This equality is not an ordinary
conditional-independence statement in $P_\eta(A,B,C,D)$. A simulator that
generates data from this latent structure therefore produces distributions
that satisfy the same equality constraint. A simulator over ordinary DAGs on
the observed variables would not generally impose
\eqref{eq:verma-running-example}. The Verma constraint and its generalizations
in nested Markov models are examples of such non-CI constraints
\cite{verma2022equivalence,shpitser2008dormant,shpitser2014nested,richardson2023nested}.

A predictor trained on datasets that satisfy these equality constraints may
use information beyond observed CI relations. In nested Markov models, some
of these constraints can be expressed as conditional independencies in a
kernel obtained by fixing. Establishing whether a particular SCD predictor
learns such constraints requires empirical evidence.

\parh{Prediction within the simulator's model class.}~A supervised predictor
can approximate an inference rule within a simulator's model class and apply
it to new datasets.
For a structural label $\tau$, the relevant identifiability condition is
\begin{equation}
\begin{gathered}
 P_\eta^{\mathrm{obs}}=P_{\eta'}^{\mathrm{obs}}
 \quad\Longrightarrow\quad \tau(\eta)=\tau(\eta'),\\
 \eta,\eta'\in\operatorname{supp}(\Pi_{\mathrm{sim}}).
\end{gathered}
\label{eq:sim-identifiability}
\end{equation}
When this condition holds, a predictor can in principle learn to estimate the
target within that model class. Finite data, model capacity, and optimization
still limit its accuracy. If the condition fails, a probabilistic predictor
can represent uncertainty among compatible labels, with their probabilities
depending on the prior. Identifiability in a different model class must be
established under that class's assumptions.

\parh{Examining simulator assumptions with SCD.}~As discussed in
\S~\ref{sec:intro}, varying the simulator can help assess which restrictions
affect a predictor. Consider a simulator for discrete data that excludes forks.
For the
three-node CPDAG $X-Y-Z$, this restriction excludes $X\leftarrow Y\to Z$
but leaves both $X\to Y\to Z$ and $X\leftarrow Y\leftarrow Z$ compatible
with the same CI relations. Restricting graph support can thus remove
representatives without identifying all remaining directions.

To examine this effect, a study can compare predicted edge probabilities at
different sample sizes, vary the prior probabilities of compatible graphs,
and evaluate data-generating models with the same observed distribution but
different graph labels.
Persistent uncertainty may indicate ambiguity, weak signal, or a limitation
of the predictor. Conversely, probabilities near zero or one may reflect a
concentrated prior, missing training cases, or poor calibration. Even a
calibrated predictor can be confident because one compatible label dominates
its training distribution. These experiments can help identify restrictions
whose effect on identifiability warrants theoretical analysis. They do not
prove the population implication in \eqref{eq:sim-identifiability} or
identify a unique graph outside the simulator family.

\subsection{Transfer, Test-time Adaptation, and Local Simulators}

A predictor trained under $\Pi_{\mathrm{sim}}$ is evaluated on tasks drawn from a
possibly different distribution $\Pi_{\mathrm{test}}$
(\F~\ref{fig:simulator-object}, bottom). With a common label rule $\tau$, its
expected test risk is
\begin{equation}
 R_{\mathrm{test}}(\theta)=
 \mathbb E_{\eta\sim\Pi_{\mathrm{test}}}
 \mathbb E_{D\sim P_\eta^{(n)}}
 \left[\ell\!\left(F_\theta(D),\tau(\eta)\right)\right].
 \label{eq:test-risk}
\end{equation}
Test risk can be high even when training risk is low because the simulator
may favor a particular ordering, omit graph patterns, or restrict mechanisms
and noise distributions. A predictor may then use features that predict
training labels accurately but are unreliable on test data. Even if training
and test distributions share causal constraints, task frequencies,
finite-sample signal strength, and the learned representation can affect
test risk.

The simulator choices in \T~\ref{tab:simulator-instantiations} illustrate
these differences. A simulator with latent variables can impose equalities
that do not hold under a different latent structure. A simulator with
interventions can provide orientation information absent from observational
test data. A restricted graph or noise prior
can also improve average prediction within the training family without
identifying the target in a broader class. These effects should be assessed
by varying the relevant simulator factors. A single measure of distributional
similarity does not distinguish these effects.

\parh{Local simulators.}~A method can use the test dataset
$D_{\mathrm{test}}$ to construct training tasks for that instance. We write
$\Pi_{\mathrm{sim}}^{D_{\mathrm{test}}}$ for the resulting local task
distribution. ML4S~\cite{ma2022ml4s} is an earlier instance: it fits a
pseudo Bayesian network and generates training data from vicinal graphs.
TTT-SCL~\cite{deng2026ttt} uses an alignment criterion
and sparsity term to refine candidate graphs in the observational setting.
TICL~\cite{chen2026ticl} uses self-augmentation with interventional data and
JCI information. These methods incur additional computation to generate
training data and fit a predictor for the test dataset.

Test-time adaptation can reduce some mismatch in observable features, but matching
an observed marginal cannot distinguish causal models that generate that
same marginal. Fitting the simulator to one dataset can also reproduce sampling noise.
Evaluation should therefore measure the benefit on held-out tasks and include
the time spent constructing tasks and retraining. A pretrained model followed
by test-time adaptation is another possible schedule. The extent to which these
procedures improve recovery on real data remains an empirical question.

\subsection{Evaluation and Trustworthy Benchmarks}

\parh{Implications for evaluation.}~A study can make its training task
reproducible by describing the simulator and label rule, including which
factors are fixed, which vary, and how they depend on one another. For example,
mechanism parameters depend on the sampled graph, and knowledge can be sampled
conditional on its ancestral relations. These factors need not be independent.
The report should also explain any conversion from predicted probabilities
to the graph being scored.

Controlled changes to graph families, mechanisms, noise, latent confounding,
interventions, or knowledge help assess specific transfer claims. The useful
choices depend on the intended deployment setting. For example, a claim of
robustness to new noise distributions calls for tests that vary those distributions.
Such evaluations provide evidence about the tested
shifts, without certifying performance under all possible shifts.

\parh{Causal reference graphs.}\label{sec:trustworthy-benchmarks}~
Held-out tasks from the training simulator test generalization to new draws
from that distribution. This is useful evidence of estimation and amortized
inference within the family. Transfer to other families or real systems
requires separate evidence. Describing the training and test simulators
allows readers to distinguish these claims.

Benchmarks differ in how their causal reference graphs are obtained.

\parh{Synthetic benchmarks.}~These benchmarks generate data from a known SCM,
so the generating graph is known. The test simulator may use the same
assumptions as the training simulator or different ones. ER/SF graphs with
linear or neural mechanisms are typical examples.

\parh{Semi-synthetic benchmarks.}~These benchmarks, also called pseudo-real
benchmarks, use mechanisms or graphs derived from a real domain but simulate
the observations. Examples include gene-expression data generated by
SERGIO~\cite{dibaeinia2020sergio} and data sampled from domain-derived networks
in the bnlearn Bayesian Network Repository~\cite{bnlearnrepository}. Their
reference graphs are specified by experts or models and need not have been
established through interventions.

\parh{Real-world benchmarks.}~These benchmarks use external causal evidence,
such as randomized experiments, interventions, or established domain
mechanisms, to construct a reference graph. The protein-signaling network of
Sachs et al.~\cite{sachs2005causal} is a widely used example. Its consensus
graph is based on biological and interventional evidence. Any uncertainty
in that graph should be considered when scoring predictions.

For each benchmark, a study should state the source of its causal reference
graph, the type of graph it represents, and any uncertain edges or
orientations. The metric should match the supported target. For example,
CPDAG evaluation avoids penalizing a method for leaving an observationally
unresolved direction open. Describing training and test simulators
separately also makes clear which results test new draws within a family and
which test transfer beyond it.

Evaluations of the compared methods illustrate these different evidence sources.
ML4S evaluates on observations sampled from repository networks, while AVICI
uses synthetic SCMs and simulated gene expression~\cite{ma2022ml4s,lorch2022amortized}.
SiCL and Arrow also evaluate on measurements from the Sachs
system~\cite{zhanglearning,thompson2026arrow}. These results do not by
themselves establish reliability in other real systems. Further benchmarks
should document the interventions or domain evidence used to construct their
reference graphs, specify uncertain edges, and use metrics matched to the
graph target.

\section{Conclusion and Future Work}
\label{sec:conclusion}

Supervised causal discovery trains a predictor on datasets with known
structural labels and applies it to a new dataset. This paper relates the
predictions of these methods to their training assumptions and the evidence
available at deployment. When several causal graphs generate the same observed
distribution, supervised learning alone cannot determine which graph generated
the data. Additional restrictions on mechanisms or noise, interventions, or
background knowledge may resolve this ambiguity. A decoder can ensure that
its output is acyclic, but identifying its edge directions still requires
these statistical assumptions or additional evidence.

The methods compared here differ in their prediction targets, model components,
and training regimes. Local classifiers estimate adjacencies or colliders,
while other predictors estimate complete graphs or posterior distributions.
Some methods also use background knowledge or estimates from classical
discovery algorithms as inputs. Simulators provide the datasets and structural
labels needed for training. Their restrictions on graphs, mechanisms, and
noise can provide information beyond conditional independence. A predictor's
encoder must retain that information for it to be useful. Test-time
adaptation methods such as ML4S also fit their training generator to the input data,
while hybrids such as SPOT retain a graph-optimization stage at deployment.
Accuracy within one simulator family does not
establish accuracy on data generated by other families.

Evaluation should therefore specify both the graph target and the training
and test assumptions. For CPDAG recovery, the metric should score the
equivalence class. For full-DAG recovery, the study should explain which
assumptions or evidence identify the additional directions. It should also
describe how model outputs are converted into the graph being scored.
These recommendations follow from the comparisons in
Sections~\ref{sec:implications-learning}--\ref{sec:simulators}.

\parh{Transfer to new data.}~Further work is needed to establish how well
supervised predictors handle mechanisms, graph structures, or latent
confounders absent from training. Broad pretraining and test-time adaptation may
help, but their benefits need to be measured on data outside the training
distribution. Evaluations of test-time adaptation should also account for the
additional computation and the risk of fitting sampling noise.

\parh{Evaluation on real data.}~Causal reference graphs for real systems are
often incomplete or disputed. Evaluating predictions against these graphs
requires documentation of the interventional or domain evidence behind them,
as well as any uncertain edges or directions. Results on the benchmarks
discussed here do not by themselves establish reliability in other real-world
settings.
Benchmarks with documented causal evidence and metrics matched to the graph
target are needed to assess progress.

\bibliographystyle{IEEEtran}
\bibliography{main}

@article{lewis1973causation,
  title={Causation},
  author={Lewis, David},
  journal={The Journal of Philosophy},
  volume={70},
  number={17},
  pages={556--567},
  year={1973},
  doi={10.2307/2025310}
}

@inproceedings{lopez2015towards,
  title={Towards a learning theory of cause-effect inference},
  author={Lopez-Paz, David and Muandet, Krikamol and Sch{\"o}lkopf, Bernhard and Tolstikhin, Iliya},
  booktitle={International Conference on Machine Learning},
  pages={1452--1461},
  year={2015},
  organization={PMLR}
}

@article{li2020supervised,
  title={Supervised Whole {DAG} Causal Discovery},
  author={Li, Hebi and Xiao, Qi and Tian, Jin},
  journal={arXiv preprint arXiv:2006.04697},
  year={2020}
}

@article{montagna2024demystifying,
  title={Demystifying Amortized Causal Discovery with Transformers},
  author={Montagna, Francesco and Cairney-Leeming, Max and Sridhar, Dhanya and Locatello, Francesco},
  journal={Transactions on Machine Learning Research},
  year={2025}
}

@inproceedings{dhirmeta,
  title={A Meta-Learning Approach to Bayesian Causal Discovery},
  author={Dhir, Anish and Ashman, Matthew and Requeima, James and van der Wilk, Mark},
  booktitle={The Thirteenth International Conference on Learning Representations},
  year={2025}
}

@article{thompson2026arrow,
  title={Arrow: A Foundation Model for Causal Discovery},
  author={Thompson, Ryan and Zhao, He and Steinberg, Daniel M. and Bonilla, Edwin V.},
  journal={arXiv preprint arXiv:2605.07204},
  year={2026}
}

@article{xu2026knowledge,
  title={A Knowledge-Informed Pretrained Model for Causal Discovery},
  author={Xu, Wenbo and He, Yue and Wang, Yunhai and Zhang, Xingxuan and Kuang, Kun and Chen, Yueguo and Cui, Peng},
  journal={arXiv preprint arXiv:2603.20842},
  year={2026}
}

@article{yin2025learning,
  title={Learning Causal Graphs at Scale: A Foundation Model Approach},
  author={Yin, Naiyu and Gao, Tian and Yu, Yue},
  journal={arXiv preprint arXiv:2506.18285},
  year={2025}
}

@article{deng2026ttt,
  title={Test Time Training for Supervised Causal Learning},
  author={Deng, Zizhen and Zhang, Jiaru and Ding, Rui and Huang, Bojun and Wang, Jinzhuo and Fu, Qiang and Han, Shi and Zhang, Dongmei},
  journal={arXiv preprint arXiv:2605.30015},
  year={2026}
}

@article{chen2026ticl,
  title={Test-Time Learning of Causal Structure from Interventional Data},
  author={Chen, Wei and Ding, Rui and Huang, Bojun and Zhang, Yang and Fu, Qiang and Liang, Yuxuan and Shi, Han and Zhang, Dongmei},
  journal={arXiv preprint arXiv:2602.19131},
  year={2026}
}

@article{wu2025sample,
  title={Sample, Estimate, Aggregate: A Recipe for Causal Discovery Foundation Models},
  author={Wu, Menghua and Bao, Yujia and Barzilay, Regina and Jaakkola, Tommi S.},
  journal={Transactions on Machine Learning Research},
  year={2025}
}

@book{hernan2010causal,
  title={Causal Inference: What If},
  author={Hern{\'a}n, Miguel A and Robins, James M},
  year={2020},
  publisher={Chapman \& Hall/CRC}
}

@book{imbens2015causal,
  title={Causal Inference for Statistics, Social, and Biomedical Sciences: An Introduction},
  author={Imbens, Guido W and Rubin, Donald B},
  year={2015},
  publisher={Cambridge University Press}
}

@article{chickering2002optimal,
  title={Optimal structure identification with greedy search},
  author={Chickering, David Maxwell},
  journal={Journal of machine learning research},
  volume={3},
  number={Nov},
  pages={507--554},
  year={2002}
}

@incollection{bareinboim2022pearl,
  title={On Pearl's Hierarchy and the Foundations of Causal Inference},
  author={Bareinboim, Elias and Correa, Juan D and Ibeling, Duligur and Icard, Thomas},
  booktitle={Probabilistic and Causal Inference: The Works of Judea Pearl},
  pages={507--556},
  year={2022},
  publisher={ACM Books},
  doi={10.1145/3501714.3501743}
}

@article{glymour2019review,
  title={Review of causal discovery methods based on graphical models},
  author={Glymour, Clark and Zhang, Kun and Spirtes, Peter},
  journal={Frontiers in Genetics},
  volume={10},
  pages={524},
  year={2019},
  publisher={Frontiers Media SA}
}

@article{zheng2018dags,
  title={{DAG}s with {NO TEARS}: Continuous Optimization for Structure Learning},
  author={Zheng, Xun and Aragam, Bryon and Ravikumar, Pradeep K and Xing, Eric P},
  journal={Advances in Neural Information Processing Systems},
  volume={31},
  year={2018}
}

@incollection{verma2022equivalence,
  title={Equivalence and synthesis of causal models},
  author={Verma, Thomas S and Pearl, Judea},
  booktitle={Probabilistic and causal inference: The works of Judea Pearl},
  pages={221--236},
  year={2022}
}

@article{ng2021reliable,
  title={Reliable causal discovery with improved exact search and weaker assumptions},
  author={Ng, Ignavier and Zheng, Yujia and Zhang, Jiji and Zhang, Kun},
  journal={Advances in Neural Information Processing Systems},
  volume={34},
  pages={20308--20320},
  year={2021}
}

@article{hoyer2008nonlinear,
  title={Nonlinear causal discovery with additive noise models},
  author={Hoyer, Patrik and Janzing, Dominik and Mooij, Joris M and Peters, Jonas and Sch{\"o}lkopf, Bernhard},
  journal={Advances in Neural Information Processing Systems},
  volume={21},
  year={2008}
}

@article{peters2014causal,
  title={Causal discovery with continuous additive noise models},
  author={Peters, Jonas and Mooij, Joris M and Janzing, Dominik and Sch{\"o}lkopf, Bernhard},
  journal={The Journal of Machine Learning Research},
  volume={15},
  number={1},
  pages={2009--2053},
  year={2014},
  publisher={JMLR.org}
}

@article{shimizu2006linear,
  title={A Linear Non-Gaussian Acyclic Model for Causal Discovery},
  author={Shimizu, Shohei and Hoyer, Patrik O and Hyv{\"a}rinen, Aapo and Kerminen, Antti and Jordan, Michael},
  journal={Journal of Machine Learning Research},
  volume={7},
  pages={2003--2030},
  year={2006}
}

@article{shimizu2011directlingam,
  title={Direct{LiNGAM}: A Direct Method for Learning a Linear Non-Gaussian Structural Equation Model},
  author={Shimizu, Shohei and Inazumi, Takanori and Sogawa, Yasuhiro and Hyv{\"a}rinen, Aapo and Kawahara, Yoshinobu and Washio, Takashi and Hoyer, Patrik O and Bollen, Kenneth},
  journal={Journal of Machine Learning Research},
  volume={12},
  pages={1225--1248},
  year={2011}
}

@inproceedings{peters2011identifiability,
  title={Identifiability of causal graphs using functional Models},
  author={Peters, Jonas and Mooij, Joris M and Janzing, Dominik and Sch{\"o}lkopf, Bernhard},
  booktitle={Proceedings of the Twenty-Seventh Conference on Uncertainty in Artificial Intelligence},
  pages={589--598},
  year={2011}
}

@inproceedings{zhang2009identifiability,
  title={On the identifiability of the post-nonlinear causal model},
  author={Zhang, Kun and Hyv{\"a}rinen, Aapo},
  booktitle={Proceedings of the Twenty-Fifth Conference on Uncertainty in Artificial Intelligence},
  pages={647--655},
  year={2009}
}

@book{pearl2009causality,
  title={Causality: Models, Reasoning, and Inference},
  author={Pearl, Judea},
  year={2009},
  publisher={Cambridge University Press},
  edition={2}
}

@inproceedings{ma2024scalable,
  title={Scalable Differentiable Causal Discovery in the Presence of Latent Confounders with Skeleton Posterior},
  author={Ma, Pingchuan and Ding, Rui and Fu, Qiang and Zhang, Jiaru and Wang, Shuai and Han, Shi and Zhang, Dongmei},
  booktitle={Proceedings of the 30th ACM SIGKDD Conference on Knowledge Discovery and Data Mining},
  pages={2141--2152},
  year={2024}
}

@inproceedings{zhanglearning,
  title={Learning Identifiable Structures Helps Avoid Bias in DNN-based Supervised Causal Learning},
  author={Zhang, Jiaru and Ding, Rui and Fu, Qiang and Huang, Bojun and Deng, Zizhen and Hua, Yang and Guan, Haibing and Han, Shi and Zhang, Dongmei},
  booktitle={Proceedings of The 28th International Conference on Artificial Intelligence and Statistics},
  pages={577--585},
  year={2025},
  volume={258},
  series={Proceedings of Machine Learning Research},
  publisher={PMLR}
}

@inproceedings{yu2019dag,
  title={{DAG-GNN}: {DAG} Structure Learning with Graph Neural Networks},
  author={Yu, Yue and Chen, Jie and Gao, Tian and Yu, Mo},
  booktitle={International Conference on Machine Learning},
  pages={7154--7163},
  year={2019},
  organization={PMLR}
}

@article{wei2020dags,
  title={{DAG}s with No Fears: A Closer Look at Continuous Optimization for Learning Bayesian Networks},
  author={Wei, Dennis and Gao, Tian and Yu, Yue},
  journal={Advances in Neural Information Processing Systems},
  volume={33},
  pages={3895--3906},
  year={2020}
}

@inproceedings{zheng2020learning,
  title={Learning Sparse Nonparametric {DAG}s},
  author={Zheng, Xun and Dan, Chen and Aragam, Bryon and Ravikumar, Pradeep and Xing, Eric},
  booktitle={International Conference on Artificial Intelligence and Statistics},
  pages={3414--3425},
  year={2020},
  organization={PMLR}
}

@article{bello2022dagma,
  title={{DAGMA}: Learning {DAG}s via {M}-matrices and a Log-Determinant Acyclicity Characterization},
  author={Bello, Kevin and Aragam, Bryon and Ravikumar, Pradeep},
  journal={Advances in Neural Information Processing Systems},
  volume={35},
  pages={8226--8239},
  year={2022}
}

@article{richardson2002ancestral,
  title={Ancestral graph Markov models},
  author={Richardson, Thomas and Spirtes, Peter},
  journal={The Annals of Statistics},
  volume={30},
  number={4},
  pages={962--1030},
  year={2002},
  publisher={Institute of Mathematical Statistics}
}

@inproceedings{shpitser2008dormant,
  title={Dormant Independence},
  author={Shpitser, Ilya and Pearl, Judea},
  booktitle={Proceedings of the Twenty-Third AAAI Conference on Artificial Intelligence},
  pages={1081--1087},
  year={2008}
}

@article{shpitser2014nested,
  title={Introduction to Nested Markov Models},
  author={Shpitser, Ilya and Evans, Robin J. and Richardson, Thomas S. and Robins, James M.},
  journal={Behaviormetrika},
  volume={41},
  number={1},
  pages={3--39},
  year={2014}
}

@article{richardson2023nested,
  title={Nested Markov Properties for Acyclic Directed Mixed Graphs},
  author={Richardson, Thomas S. and Evans, Robin J. and Robins, James M. and Shpitser, Ilya},
  journal={The Annals of Statistics},
  volume={51},
  number={1},
  pages={334--361},
  year={2023}
}

@article{vowels2022d,
  title={D'ya Like {DAG}s? A Survey on Structure Learning and Causal Discovery},
  author={Vowels, Matthew J and Camg{\"o}z, Necati Cihan and Bowden, Richard},
  journal={ACM Computing Surveys},
  volume={55},
  number={4},
  pages={82:1--82:36},
  year={2023},
  publisher={ACM New York, NY},
  doi={10.1145/3527154}
}

@book{spirtes2000causation,
  title={Causation, Prediction, and Search},
  author={Spirtes, Peter and Glymour, Clark N and Scheines, Richard and Heckerman, David},
  year={2000},
  publisher={MIT Press},
  edition={2}
}

@article{zhang2008completeness,
  title={On the completeness of orientation rules for causal discovery in the presence of latent confounders and selection bias},
  author={Zhang, Jiji},
  journal={Artificial Intelligence},
  volume={172},
  number={16-17},
  pages={1873--1896},
  year={2008},
  publisher={Elsevier}
}

@article{tsamardinos2006max,
  title={The max-min hill-climbing Bayesian network structure learning algorithm},
  author={Tsamardinos, Ioannis and Brown, Laura E and Aliferis, Constantin F},
  journal={Machine Learning},
  volume={65},
  pages={31--78},
  year={2006},
  publisher={Springer}
}

@inproceedings{ma2022ml4s,
  title={{ML4S}: Learning Causal Skeleton from Vicinal Graphs},
  author={Ma, Pingchuan and Ding, Rui and Dai, Haoyue and Jiang, Yuanyuan and Wang, Shuai and Han, Shi and Zhang, Dongmei},
  booktitle={Proceedings of the 28th ACM SIGKDD Conference on Knowledge Discovery and Data Mining},
  pages={1213--1223},
  year={2022}
}

@book{peters2017elements,
  title={Elements of causal inference},
  author={Peters, Jonas and Janzing, Dominik and Sch{\"o}lkopf, Bernhard},
  year={2017},
  publisher={The MIT Press}
}

@inproceedings{lorch2022amortized,
  title={Amortized Inference for Causal Structure Learning},
  author={Lorch, Lars and Sussex, Scott and Rothfuss, Jonas and Krause, Andreas and Sch{\"o}lkopf, Bernhard},
  booktitle={Advances in Neural Information Processing Systems},
  volume={35},
  pages={13104--13118},
  year={2022}
}

@inproceedings{dai2023ml4c,
  title={{ML4C}: Seeing Causality through Latent Vicinity},
  author={Dai, Haoyue and Ding, Rui and Jiang, Yuanyuan and Han, Shi and Zhang, Dongmei},
  booktitle={Proceedings of the 2023 SIAM International Conference on Data Mining (SDM)},
  pages={226--234},
  year={2023},
  organization={SIAM}
}

@inproceedings{ke2022learning,
  title={Learning to Induce Causal Structure},
  author={Ke, Nan Rosemary and Chiappa, Silvia and Wang, Jane X and Bornschein, Jorg and Goyal, Anirudh and Rey, Melanie and Weber, Theophane and Botvinick, Matthew and Mozer, Michael Curtis and Rezende, Danilo Jimenez},
  booktitle={International Conference on Learning Representations},
  year={2023}
}

@article{colombo2012learning,
  title={Learning high-dimensional directed acyclic graphs with latent and selection variables},
  author={Colombo, Diego and Maathuis, Marloes H and Kalisch, Markus and Richardson, Thomas S},
  journal={The Annals of Statistics},
  volume={40},
  number={1},
  pages={294--321},
  year={2012}
}

@article{sachs2005causal,
  title={Causal protein-signaling networks derived from multiparameter single-cell data},
  author={Sachs, Karen and Perez, Omar and Pe'er, Dana and Lauffenburger, Douglas A and Nolan, Garry P},
  journal={Science},
  volume={308},
  number={5721},
  pages={523--529},
  year={2005},
  doi={10.1126/science.1105809}
}

@inproceedings{ogarrio2016hybrid,
  title={A hybrid causal search algorithm for latent variable models},
  author={Ogarrio, Juan Miguel and Spirtes, Peter and Ramsey, Joe},
  booktitle={Conference on Probabilistic Graphical Models},
  pages={368--379},
  year={2016},
  organization={PMLR}
}

@inproceedings{meek1995causal,
  title={Causal inference and causal explanation with background knowledge},
  author={Meek, Christopher},
  booktitle={Proceedings of the Eleventh Conference on Uncertainty in Artificial Intelligence},
  pages={403--410},
  year={1995}
}

@inproceedings{zhang2011kernel,
  title={Kernel-based conditional independence test and application in causal discovery},
  author={Zhang, Kun and Peters, Jonas and Janzing, Dominik and Sch{\"o}lkopf, Bernhard},
  booktitle={Proceedings of the Twenty-Seventh Conference on Uncertainty in Artificial Intelligence},
  pages={804--813},
  year={2011}
}

@inproceedings{nazaret2024stable,
  title={Stable differentiable causal discovery},
  author={Nazaret, Achille and Hong, Justin and Azizi, Elham and Blei, David},
  booktitle={Proceedings of the 41st International Conference on Machine Learning},
  pages={37413--37445},
  year={2024}
}

@article{halpern2005causes,
  title={Causes and Explanations: A Structural-Model Approach. Part I: Causes},
  author={Halpern, Joseph Y. and Pearl, Judea},
  journal={The British Journal for the Philosophy of Science},
  volume={56},
  number={4},
  pages={843--887},
  year={2005},
  doi={10.1093/bjps/axi147}
}

@article{dibaeinia2020sergio,
  title={{SERGIO}: A Single-Cell Expression Simulator Guided by Gene Regulatory Networks},
  author={Dibaeinia, Payam and Sinha, Saurabh},
  journal={Cell Systems},
  volume={11},
  number={3},
  pages={252--271},
  year={2020}
}

@misc{bnlearnrepository,
  author={Scutari, Marco},
  title={Bayesian Network Repository},
  howpublished={\url{https://www.bnlearn.com/bnrepository/}}
}

\end{document}